\documentclass[lettersize,journal]{IEEEtran}

\usepackage{subfiles}
\usepackage{titling}  

\usepackage{amsfonts}
\usepackage{algorithmic}
\usepackage{algorithm}
\usepackage{array}
\usepackage[caption=false,font=normalsize,labelfont=sf,textfont=sf]{subfig}
\usepackage{textcomp}
\usepackage{stfloats}
\usepackage{url}
\usepackage{verbatim}
\usepackage{graphicx}
\usepackage{cite}
\usepackage{siunitx}
\usepackage{gensymb}
\usepackage[hidelinks]{hyperref} 
\usepackage[nolist]{acronym}
\usepackage{booktabs} 
\usepackage[numbers]{natbib}
\usepackage{tabularx}
\usepackage{multirow}
\usepackage[dvipsnames]{xcolor}
\usepackage[fleqn]{amsmath} 
\usepackage{tikz}
\usetikzlibrary{positioning}
\usepackage{verbatim}
\usepackage{pgfplots}
\usepackage{array}
\usepackage{booktabs} 
\definecolor{bleudefrance}{rgb}{0.19, 0.55, 0.91}
\definecolor{lava}{rgb}{0.97, 0.41, 0.02}
\definecolor{bluegray}{rgb}{0.4, 0.6, 0.8}
\definecolor{blue(ncs)}{rgb}{0.0, 0.53, 0.74}
\definecolor{brandeisblue}{rgb}{0.0, 0.44, 1.0}
\definecolor{coolblack}{rgb}{0.0, 0.18, 0.39}
\newcommand{\uproman}[1]{\uppercase\expandafter{\romannumeral#1}} 
\usepackage{pdfpages}

\begin{acronym}
	\acro{SimMIM}[SimMIM]{Simple Framework for Masked Image Modeling}
	\acro{resnet}[ResNet]{Residual Network}
	\acro{aspp}[ASPP]{Atrous Spatial Pyramid Pooling}
	\acro{mde}[MDE]{Mean Distance Error}
	\acro{caffe}[CaFFe]{``CAlving Fronts and where to Find thEm''}
	\acro{sar}[SAR]{Synthetic Aperture Radar}
	\acro{SAM}[SAM]{Segment Anything Model}
	\acro{NA}[NA]{``no information available''}
	\acro{ViT}[ViT]{Vision Transformer}
	\acro{crf}[CRF]{Conditional Random Field}
	\acro{CNN}[CNN]{Convolutional Neural Network}
	\acro{AMD-HookNet}[AMD-HookNet]{attention-multihooking-deep-supervision HookNet}
	\acro{crf}[CRF]{Conditional Random Field}
	\acro{gee}[GEE]{Google Earth Engine}
	\acro{sam}[SAM]{Segment Anything Model}
	\acro{tsx}[TSX]{TerraSAR-X}
	\acro{tdx}[TDX]{TanDEM-X}
	\acro{wmo}[WMO]{World Meteorological Organization}
	\acro{ecv}[ECV]{Essential Climate Variable}
	\acro{gla-st}[GLA-ST]{global-local attention Swin-Transformer block}
	\acro{gla}[GLA-STDeepLab]{global-local attention Swin-Transformer-based DeepLabv3+}
	\acro{mtl}[MTL]{multi-task learning}
	\acro{aspp}[ASPP]{Atrous Spatial Pyramid Pooling}
	\acro{dl}[DL]{Deep Learning}
	\acro{mcc}[MCC]{Matthew's Correlation Coefficient}
	\acro{ssl}[SSL]{self-supervised learning}
	\acro{gan}[GAN]{generative adversarial network}
	\acro{mim}[MIM]{masked image modeling}
	\acro{s1}[S1]{Sentinel-1}
	\acro{s2}[S2]{Sentinel-2}
	\acro{aot}[AOT]{Aerosol Optical Thickness}
	\acro{wvp}[WVP]{Average Water Vapor}
	\acro{scl}[SCL]{Scene Classification Layer}
	\acro{tci}[TCI]{true-color image}
	\acro{cldprb}[MSK\_CLDPRB]{Cloud Probability Mask}
	\acro{snwprb}[MSK\_SNWPRB]{Snow Probability Mask}
	\acro{qa10}[QA10]{Cloud Mask}
	\acro{gee}[GEE]{Google Earth Engine}
	\acro{grd}[GRD]{Ground Range Detected}
\end{acronym}

\begin{document}

\setlength{\mathindent}{0cm}
\definecolor{bluegray}{rgb}{0.4, 0.6, 0.8}
\definecolor{blue(ncs)}{rgb}{0.0, 0.53, 0.74}
\definecolor{brandeisblue}{rgb}{0.0, 0.44, 1.0}
\definecolor{coolblack}{rgb}{0.0, 0.18, 0.39}

\title{SSL4SAR: Self-Supervised Learning for Glacier Calving Front Extraction from SAR Imagery}

\author{Nora Gourmelon$^{1, *}$, Marcel Dreier$^1$, Martin Mayr$^2$, Thorsten Seehaus$^3$, Dakota Pyles$^3$, Matthias Braun$^3$, \\Andreas Maier$^1$, Vincent Christlein$^1$
\thanks{$^*$ Corresponding author, nora.gourmelon@fau.de\\
$^1$Pattern Recognition Lab, Computer Science Department,
Friedrich-Alexander-Universität Erlangen-Nürnberg, Erlangen, Germany\\
$^2$Erlangen National High Performance Computing Center (NHR@FAU),
Friedrich-Alexander-Universität Erlangen-Nürnberg,
Erlangen, Germany\\
$^3$Institut für Geographie, Department of Geography and Geosciences,
Friedrich-Alexander-Universität Erlangen-Nürnberg, Erlangen, Germany\\
}
\thanks{© 2025 IEEE.  Personal use of this material is permitted.  Permission from IEEE must be obtained for all other uses, in any current or future media, including reprinting/republishing this material for advertising or promotional purposes, creating new collective works, for resale or redistribution to servers or lists, or reuse of any copyrighted component of this work in other works.}
}
\predate{}
\postdate{}
\date{}

\maketitle

\begin{abstract}
Glaciers are losing ice mass at unprecedented rates, increasing the need for accurate, year-round monitoring to understand frontal ablation, particularly the factors driving the calving process. Deep learning models can extract calving front positions from Synthetic Aperture Radar imagery to track seasonal ice losses at the calving fronts of marine- and lake-terminating glaciers.
The current state-of-the-art model relies on ImageNet-pretrained weights. 
However, they are suboptimal due to the domain shift between the natural images in ImageNet and the specialized characteristics of remote sensing imagery, in particular for Synthetic Aperture Radar imagery.
To address this challenge, we propose two novel self-supervised multimodal pretraining techniques that leverage SSL4SAR, a new unlabeled dataset comprising 9,563 Sentinel-1 and 14 Sentinel-2 images of Arctic glaciers, with one optical image per glacier in the dataset. Additionally, we introduce a novel hybrid model architecture that combines a Swin Transformer encoder with a residual \ac{CNN} decoder. When pretrained on SSL4SAR, this model achieves a mean distance error of 293\,m on the \acf{caffe} benchmark dataset, outperforming the prior best model by 67\,m. 
Evaluating an ensemble of the proposed model on a multi-annotator study of the benchmark dataset reveals a mean distance error of 75\,m, approaching the human performance of 38\,m. This advancement enables precise monitoring of seasonal changes in glacier calving fronts.
\end{abstract}


\section{Introduction}

\IEEEPARstart{G}{laciers} are facing record ice losses, with the largest annual mass loss rate ever recorded at $602\,\pm\,$\SI{69}{\giga\tonne} in 2023, and the trend is upward, as global glacier mass loss has been increasing exponentially in recent years~\cite{Dussaillant.2024}.                          
For marine-terminating glaciers, which account for 40\,\% of total glacier area~\cite{RGI.2017}, ice depletion is controlled by two mechanisms: negative surface mass balance and ice losses at the calving front~\cite{Khan.2015,Sheperd.2018}.
Glacier responses to ocean forcing and ice-ocean interaction are to date not well understood~\cite{IPCC.2021}. 
To identify drivers of frontal ablation, and hence mass loss at the terminus, it is necessary to assess calving front variations.
The community lacks large spatiotemporal databases of calving front positions that can be used as a basis for further analysis.
Due to the impracticality of manually delineating calving fronts in satellite imagery, several attempts have been made in recent years to automate the extraction of front positions using deep learning (e.g.,~\cite{Heidler.2023} and~\cite{Loebel.2022}).
Since seasonal calving front dynamics are of interest, and polar night and weather conditions limit the availability of optical images, we focus on \ac{sar} images in this study. 
However, \ac{sar} images are more difficult to interpret than optical images due to speckle noise, often lower resolution, and similar backscatter patterns for different ice types~\cite{Baumhoer.2018}.
On the other hand, the contrast between the open ocean and glacier ice is high for \ac{sar} images, particularly when ice mélange and sea ice are absent at the glacier front.
For calving front delineation, a benchmark dataset with only \ac{sar} imagery exists: \acf{caffe}~\cite{Gourmelon-2022}.
A comparison of previously published deep learning models for calving front and coastline extraction on this benchmark~\cite{Gourmelon.2025} shows that the best performing model is based on a Swin-Transformer-based architecture without convolutions, denoted as ``HookFormer''~\cite{Wu.2023_2}.
The model addresses the perennial challenge for deep learning models in remote sensing: how to incorporate surrounding contextual information into the local patch, with two U-shaped branches, where one receives the local patch and the other receives a surrounding, larger, downscaled patch.
A drawback of this method is that it produces jagged calving front delineations.
Therefore, this study proposes an optimized, hybrid transformer-\ac{CNN} architecture that reduces model size by eliminating the two-branch system and produces smooth calving front delineations through its \ac{CNN} decoder.

HookFormer is currently the only model for calving front delineation that relies on pretrained weights rather than random initialization. Specifically, it uses Swin-Transformer~\cite{Liu.2021} weights trained on the ImageNet dataset~\cite{Deng.2009}, which are publicly available\footnote{\url{https://github.com/microsoft/Swin-Transformer}}. This pretraining was crucial for achieving state-of-the-art performance and highlights the importance of using pretrained models (see Tab.~\ref{tab:distance_errors_sensors_mde}).
ImageNet consists of natural images that have specific properties which distinguish them from remote sensing images, such as perspective, scale, and spatial resolution.
In addition, the images in ImageNet are optical images, unlike the \ac{sar} images in \ac{caffe}.
Such a discrepancy in the domain between the source dataset used for pretraining and the target dataset used for fine-tuning can affect the performance of the model in the downstream task~\cite{Ren.2024, Dimitrovski.2024, Wang.2022, Manas.2021, Li.2023}.
To leverage and further expand the performance gains of pretraining for calving front delineation, we introduce a new unlabeled \ac{sar} dataset of calving glaciers together with two novel in-domain, multimodal pretraining strategies that use \ac{sar} images as input and optical images as labels.

Finally, we present a new state-of-the-art deep learning ensemble model for calving front delineation in \ac{sar} images, which achieves near-human performance in a comparison with the multi-annotator study on the \ac{caffe} benchmark~\cite{Gourmelon.2025}. 
Our contributions can be summarized as follows:
\begin{itemize}
    \item Introduction of a \ac{sar} dataset of arctic glaciers for self-supervised pretraining.
    \item Two novel multimodal pretraining strategies.
    \item Proposal of an optimized single-branch model architecture.
    \item New state-of-the-art ensemble model for calving front delineation in \ac{sar} imagery with near human performance.
\end{itemize}\

The remainder of the paper is organized as follows.
The next section (Sect.~\ref{sec:related_work}) gives an overview of the related work.
Sect.~\ref{sec:datasets} describes the datasets used.
The proposed architecture and pretraining strategies are explained in Sect.~\ref{sec:methodology}, while Sect.~\ref{sec:experiments} outlines the evaluation methods, training setup and ensemble approach.
The results are presented in Section~\ref{sec:results}, followed by an ablation study on the ensemble approach components in Section~\ref{sec:ablation}.
Finally, Sections~\ref{sec:discussion} and~\ref{sec:conclusion} provide a discussion of the findings and conclude the paper.

\section{Related Work}
\label{sec:related_work}
Several studies have proposed deep learning models for extracting calving front or coastline positions from satellite imagery.
So far, only one of these studies has used supervised pretraining, and none have performed in-domain pretraining.
However, there is a large body of literature on self-supervised pretraining for remote sensing.
In the following sections, we review studies on delineating calving fronts and coastlines in satellite imagery and studies on self-supervised pretraining for remote sensing.

    \subsection{Calving Front Delineation}

    Classical calving front extraction approaches~\cite{Liu.2004, Krieger.2017, Sohn.1999, Seale.2011, Liu.2021_classical} rely on either abrupt intensity changes at the terminus, relatively homogeneous neighboring glacier ice and ocean areas with different average intensities, or both~\cite{Liu.2004}. However, these assumptions often do not hold in complex \ac{sar} scenes, limiting the generalizability of such methods. To address this, recent research has increasingly adopted data-driven techniques.
    Deep learning models for delineating calving fronts or coastlines from satellite imagery can be broadly categorized into \ac{CNN}-based and transformer-based architectures.
    
    The majority of studies rely on \acp{CNN}~\cite{Baumhoer.2019, Cheng.2021, Davari_Baller.2021, Davari_Islam.2021, Gourmelon-2022, Gourmelon.2023, Hartmann.2021, Heidler.2021, Heidler.2023, Herrmann.2023, Holzmann.2021, Loebel.2022, Marochov.2021, Mohajerani.2019, Periyasamy_2022, Wu.2023_1, Zhang.2019, Zhang.2021, Zhang.2023}.
    Since the receptive field of \acp{CNN} is locally bounded, techniques are needed to introduce global information.
    Widely used are encoder-decoder architectures such as the U-Net~\cite{Ronneberger.2015} and DeepLabv3+~\cite{Chen_2018_ECCV}, which use multiple pooling operations to aggregate information from more distant locations~\cite{Baumhoer.2019, Cheng.2021, Davari_Baller.2021, Davari_Islam.2021, Gourmelon-2022, Gourmelon.2023, Hartmann.2021, Heidler.2021, Herrmann.2023, Holzmann.2021, Loebel.2022, Mohajerani.2019, Periyasamy_2022, Wu.2023_1, Zhang.2019, Zhang.2021, Zhang.2023}.
    Deeper architectures~\cite{Loebel.2022}, i.\,e., with more pooling operations and larger kernel sizes~\cite{Zhang.2019}, are sometimes used to enhance the aggregation effect of encoder-decoder structures.
    To further increase the receptive field, several studies use \ac{aspp}~\cite{Cheng.2021, Gourmelon-2022, Gourmelon.2023, Periyasamy_2022, Zhang.2021, Zhang.2023, Zhu.2023} 
    and dilated (atrous) convolutions in general~\cite{Periyasamy_2022}. 
    Two studies~\cite{Heidler.2021, Herrmann.2023} apply deep supervision to reinforce the global information aggregated by their encoder-decoder architectures. 
    In post-processing, Gourmelon et al.~\cite{Gourmelon.2023} incorporate a \ac{crf}, which, unlike simple thresholding, considers the potentially assigned classes of all other pixels when determining the class of a given pixel based on the neural network's output.
    Finally, Wu et al.~\cite{Wu.2023_1} implement a two branch model that receives a local and a context patch.
    Information is transferred from the context branch to the local branch, effectively increasing the field of view.
    
    Transformer-based networks, on the other hand, have access to global information due to the inherent attention mechanism.
    The successor model~\cite{Wu.2023_2} of Wu et al.~\cite{Wu.2023_1}'s two branch model still has two branches, but replaces the convolutions with Swin-Transformer~\cite{Liu.2021} blocks.
    Zhu et al.~\cite{Zhu.2023} combine transformers and \acp{CNN} by incorporating global-local attention Swin-Transformer blocks in DeepLabv3+.
    
    \subsection{Self-supervised pretraining for Remote Sensing}
    With the increasing availability of ever-larger unlabeled datasets, self-supervised pretraining is on the rise and has not stopped at remote sensing.
    
    Contrastive learning is a widely used \ac{ssl} technique that reduces the distance between semantically identical samples (positive pairs) in feature space and, for some approaches, also increases the distance between samples that are not semantically identical (negative pairs)~\cite{Wang.2022_review}. 
    In remote sensing, common ways to generate positive pairs include flipping~\cite{Calhoun.2022, Dimitrovski.2024, Li.2023, Manas.2021, Muhtar.2023, Tolan.2024}, rotating~\cite{Li.2023}, cropping~\cite{Calhoun.2022, Dimitrovski.2024, Li.2023, Manas.2021, Muhtar.2023, Tolan.2024, Wang.2022, Wanyan.2024}, resizing~\cite{Dimitrovski.2024, Tolan.2024}, masking~\cite{Dimitrovski.2024, Muhtar.2023, Tolan.2024}, shifting~\cite{Chen.2022}, blurring~\cite{Calhoun.2022, Li.2023, Wang.2022, Wanyan.2024}, color jitter~\cite{Manas.2021, Wang.2022, Wanyan.2024}, grayscale shifting~\cite{Wanyan.2024}, solarization~\cite{Wang.2022, Wanyan.2024}, Gaussian noise~\cite{Wang.2022_review}, and, for multimodal models sensor dropping~\cite{Wang.2022}, and any combination of the above.
    If a multi-temporal dataset is used, positive pairs can also be formed by taking temporally close images~\cite{Akiva.2022, Manas.2021, Wanyan.2024}.
    
    In contrast, generative \ac{ssl} techniques typically reconstruct the input data or generate new data based on the input, often employing autoencoders or \acp{gan}~\cite{Wang.2022_review}.
    A classic pre-text task for generative \ac{ssl} is \ac{mim}~\cite{He.2022, Xie.2022}, which involves masking parts of the input with zeros and subsequently reconstructing them using the model.
    However, besides \ac{mim}~\cite{Xiong.2024}, pre-text tasks such as denoising~\cite{Nguyen.2021_CVPR} or band switching~\cite{Qian.2022} are also used for pretraining in remote sensing.
    
    Most remote sensing studies rely primarily on optical imagery for pretraining.
    A work that takes \ac{sar}-optical image pairs as input is presented by Scheibenreif et\,al.~\cite{Scheibenreif.2022}.
    The approach uses a model for each modality and applies a contrastive loss to bring the \ac{sar}-extracted and optical-extracted features closer together.
    Similarly, Chen et\,al.~\cite{Chen.2022} and Wang et\,al.~\cite{Wang.2022} take \ac{sar}-optical image pairs as input, but only employ one shared model with the modalities in separate channels.
    In Chen et\,al.~\cite{Chen.2022}, both the pretraining and the fine-tuning use multimodal data, whereas Wang et\,al.~\cite{Wang.2022} show that multimodal pretraining helps downstream tasks with multimodal, only \ac{sar}, and only optical data.
    A work that takes only \ac{sar} images as input is presented by Li et\,al.~\cite{Li.2023}.
    However, their contrastive approach still requires \ac{sar}-optical image pairs.
    To generate a positive sample for the real \ac{sar} image, they employ a \ac{gan} to create a fake \ac{sar} image based on the optical image.
    
    In contrast, our multimodal, generative pretraining technique does not necessitate temporally aligned \ac{sar}-optical image pairs and manages with only a limited number of geographically co-registered optical images.
    
\section{Datasets}
\label{sec:datasets}
This work introduces novel multimodal pretraining approaches that make use of new unlabeled data. 
For fine-tuning and evaluation, we use the manually labeled \ac{caffe} benchmark dataset~\cite{Gourmelon-2022-caffe}.

    \subsection{SSL4SAR}
    The unlabeled dataset for pretraining, which we call SSL4SAR, comprises 9,562 \ac{s1} images in Interferometric Wide swath mode covering the period 2015-2022, depicting 14 distinct glaciers in the Arctic with different sizes and calving front geometries.
    A bounding box around each glacier terminus was manually defined.
    In order to efficiently obtain time-series of \ac{sar} imagery for the target glacier, we relied on \ac{s1} \ac{sar} \ac{grd} image collection (ee.ImageCollection(``COPERNICUS/S1\_GRD'')) provided by the database from \ac{gee}. In contrast to the original \ac{s1} \ac{sar} \ac{grd} imagery provided by Copernicus, \ac{gee} applied additional preprocessing steps to enhance the quality of the data: 1. Apply orbit files; 2. \ac{grd} broader noise removal; 3. Thermal noise removal; 4. Radiometric calibration; 5. Terrain correction.
    At each target glacier, the \ac{s1} imagery was subsetted to the defined bounding box and reprojected to Arctic Polar Stereographic projection (EPSG: 3995) for the glaciers on Svalbard and UTM zone 6N projection (EPSG:32606) for the Columbia Glacier at \SI{10}{\metre} spatial resolution.
    
    Additionally to the SAR imagery, for each of these glaciers, one cloud-free, optical image captured by \ac{s2} between June and August 2020 is included in the dataset.
    The images were manually selected from the \ac{gee} \ac{s2} MSI collection (ee.ImageCollection(``COPERNICUS/S2\_SR'')). Similar to the \ac{s1} imagery, the multi-spectral data was cropped to the bounding boxes defined at each glacier terminus and reprojected to \SI{10}{\metre} spatial resolution at the respective projections used for the \ac{sar} imagery at each glacier.
    The multi-spectral images feature all twelve bands as well as the data products \ac{aot}, \acf{wvp}, \acf{scl}, \ac{tci}, \ac{cldprb}, \ac{snwprb}, and Cloud Masks (QA10, QA20, QA60). 
    However, we only utilize the twelve bands, \ac{wvp}, and \ac{scl} in our pretraining approaches.
    \ac{wvp} and \ac{scl} offer additional informative supervision signals, providing semantically meaningful spatial structures that can be visually interpreted and are correlated with surface characteristics relevant to the downstream task of landscape zone segmentation.
    
    \subsection{CaFFe}
    \ac{caffe} consists of 681 \ac{sar} images acquired by six different sensors with varying spatial resolutions, one of which is \ac{s1}.
    The dataset includes scenes captured from both ascending and descending orbits, resulting in a range of viewing geometries and incidence angles.
    The acquisition period runs from 1996 to 2020.
    The dataset depicts seven marine-terminating glaciers, five situated on the Antarctic Peninsula, one in Greenland, and one in Alaska.
    Following the official train-test split, all images from two of these glaciers -- Mapple and Columbia glaciers --  are assigned to the test set, resulting in a total of 122 test images.
    \ac{caffe} provides two labels for each image.
    One label shows the calving front of the glacier, while the other label, which is intended as an intermediate objective for segmentation approaches, depicts landscape zones (ocean, glacier, rock, no information available).
    For further information on \ac{caffe}, we refer the reader to~\cite{Gourmelon-2022}.
    To assess the reliability of the manual annotations and establish a human performance reference for automated calving front extraction, Gourmelon et al.~\cite{Gourmelon.2025} quantified the inter-annotator variability by computing the \ac{mde} between calving front delineations provided by ten independent annotators across all images in the test set.
    
\section{Methodology}
\label{sec:methodology}
A major challenge for deep learning models in remote sensing is the integration of context information into the model.
Previous studies~\cite{Wu.2023_1, Wu.2023_2} for calving front delineation in \ac{sar} imagery have used dual branch architectures.
With our new hybrid transformer-\ac{CNN} model, 
we eliminate the need for a second branch.

    \subsection{Architecture}
    \begin{figure*}[t]
        \centering
        \includegraphics[width=\textwidth]{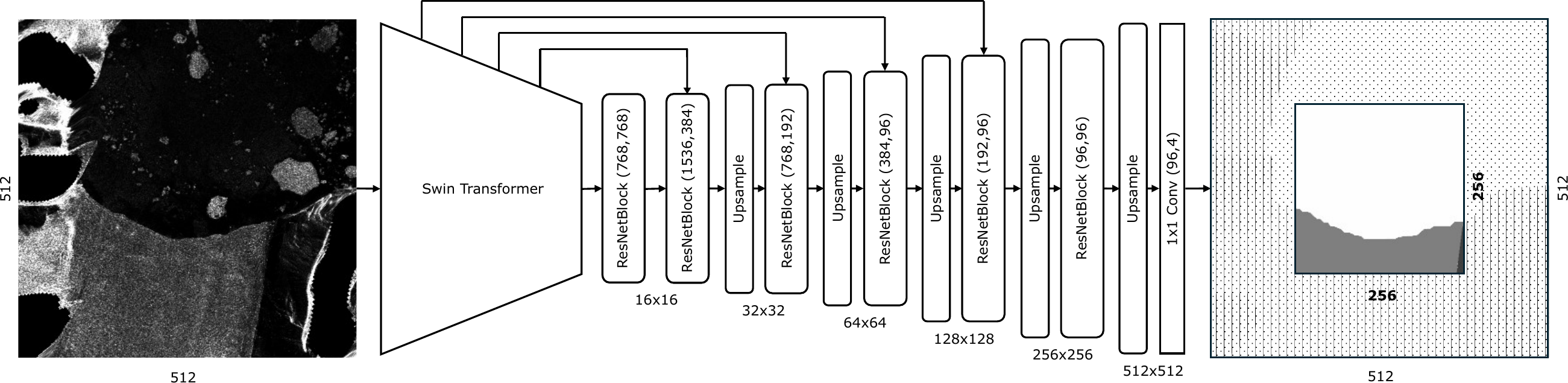}
        \caption{Architecture of the network, dubbed ``TYRION'' (\textbf{T}ransformer H\textbf{y}brid with a \textbf{R}es\textbf{i}dual C\textbf{o}nvolutional \textbf{N}eural Network). The Swin Transformer requires three input channels, so the single-channel \ac{sar} image is replicated three times before being input into the model. The output consists of four channels, each representing a class: Ocean (white), glacier (light grey), rock outcrop (dark grey), and areas with no information. The outer frame of the output is not considered during testing and inference but included during training. The \ac{sar} image is shown without contrast enhancement to reflect the raw input provided to the model.}
        \label{fig:architecture}
    \end{figure*}\
    \begin{figure}[htbp]
        \centering
        \includegraphics[width=0.35\textwidth]{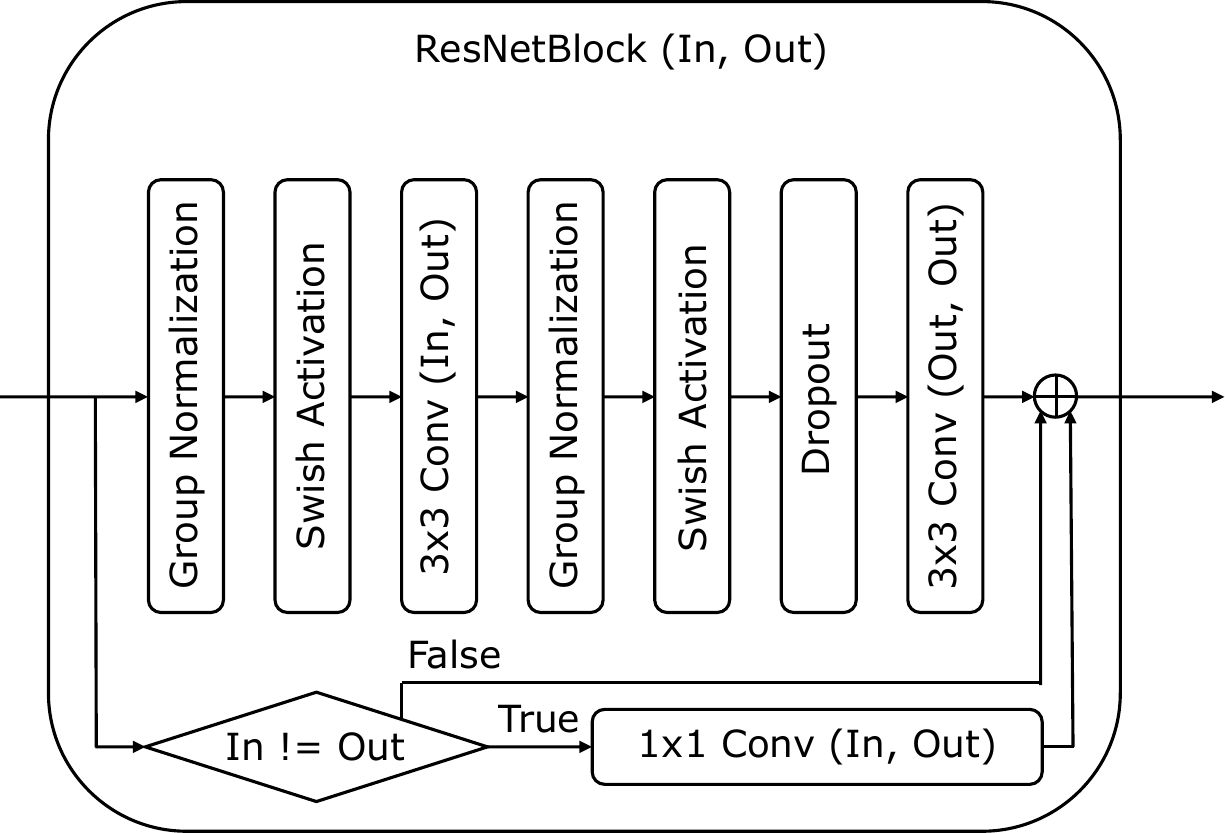}
        \caption{Design of the ResNetBlock. $\oplus$ indicates addition.}
        \label{fig:ResNetBlock}
    \end{figure}\
    We take inspiration from the current best performing method, the fully Transformer-based model ``HookFormer''~\cite{Wu.2023_2}. 
    However, we simplify its architecture and overcome the problem of jagged fronts produced by the original model.
    The model receives a single-channel \ac{sar} image, replicated three times to match the encoder's input channel count, and outputs four segmentation maps -- one for each landscape zone.
    The model's architecture is illustrated in Fig.~\ref{fig:architecture}.
    The encoder-decoder structure is a hybrid between a transformer and a \ac{CNN}:
    The network's encoder is a SwinV2-Transformer~\cite{Liu.2022} with four downsampling steps and window size of $16$, while the \ac{CNN} decoder consists of residual blocks~\cite{He_2016_CVPR} and upsampling steps. The design of the residual block, taken from~\cite{Esser.2021}, is depicted in Fig.~\ref{fig:ResNetBlock} and the upsampling is performed by nearest neighbor interpolation.
    The \ac{CNN} decoder produces zone segmentations that result in smooth fronts after post-processing.
    To eliminate the need of a second branch, we increase the input size of the model from $224 \times 224$ to $512 \times 512$. 
    The input size is maintained by the network, but we retain only the inner $256 \times 256$ patch to ensure that a sufficient amount of surrounding context information is used for the output. 
    We dub our network ``TYRION'' (\textbf{T}ransformer H\textbf{y}brid with a \textbf{R}es\textbf{i}dual C\textbf{o}nvolutional \textbf{N}eural Network).
            
    \subsection{SSL Pretraining}
    When humans delineate calving fronts on \ac{sar} imagery, they often use an additional optical image of the glacier as a reference to support the mapping.
    Deep learning models can also benefit from multimodal training, as useful and possibly complementary information is transferred from both modalities, thus improving the quality of latent spatial features~\cite{huang.2021, Wang.2023_SSL4EO}. 
    In line with the typical human approach, we have implemented two multimodal pretraining techniques: ``SSL4SAR\,--\,OptSimMIM'' (Optical supervised SimMIM trained on SSL4SAR) and ``SSL4SAR\,--\,OptTranslator'' (\ac{sar} to Optical Translator trained on SSL4SAR).
    In contrast to previous studies~\cite{Scheibenreif.2022, Chen.2022, Wang.2022, Li.2023}, both presented pretraining techniques use only one additional optical image per glacier, rather than a temporally matching optical image for each \ac{sar} image in the SSL4SAR dataset.
    This markedly reduces the need for manual data curation, as the number of cloud-free matching optical images to be collected is reduced from 9,562 to 14 in the case of SSL4SAR.

        \subsubsection{SSL4SAR\,--\,OptSimMIM}
\begin{figure}[ht]
            \centering
            \includegraphics[width=0.48\textwidth]{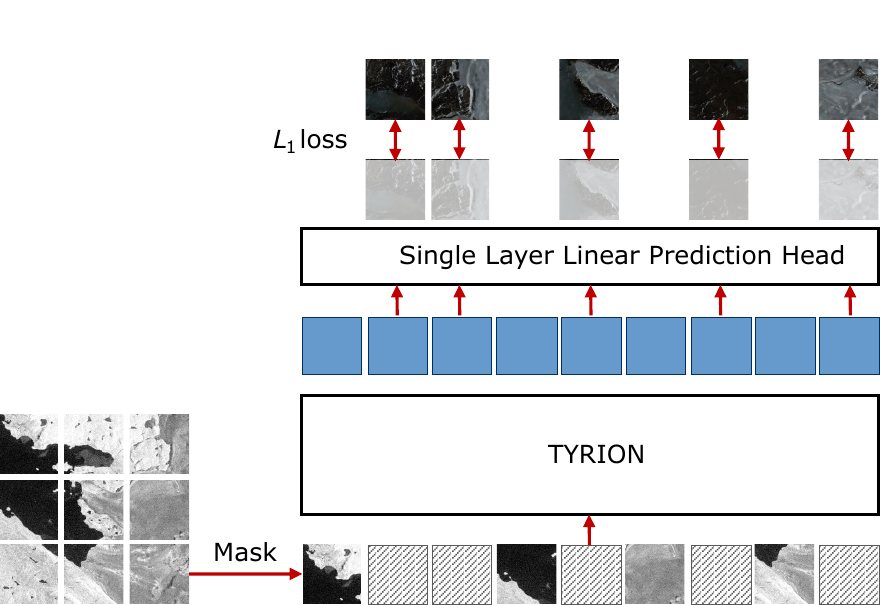}
            \caption{Visualization of SSL4SAR\,--\,OptSimMIM. For illustration purposes, we show only the rgb channels of the optical image. The striped patches are overwritten with mask tokens consisting of zeros. The \ac{sar} image is shown without contrast enhancement to reflect the raw input provided to the model.} Figure adapted from~\cite{Xie.2022}.
            \label{fig:SimMIM}
        \end{figure}

        The first pretraining approach, called SSL4SAR\,--\,OptSimMIM, leverages the \acf{SimMIM}~\cite{Xie.2022}.
        SimMIM belongs to the generative \ac{ssl} methods.
        It masks random patches of the input image and uses a one-layer prediction head to predict the raw pixel values of these masked patches.
        SSL4SAR\,--\,OptSimMIM receives a \ac{sar} image and uses the one-layer prediction head to predict the values of the corresponding patches of the optical image from the same glacier instead of predicting the \ac{sar} patches.
        A visualization of SSL4SAR\,--\,OptSimMIM is given in Fig.~\ref{fig:SimMIM}.

        \subsubsection{SSL4SAR\,--\,OptTranslator}
\begin{figure}[ht]
            \centering
            \includegraphics[width=0.48\textwidth]{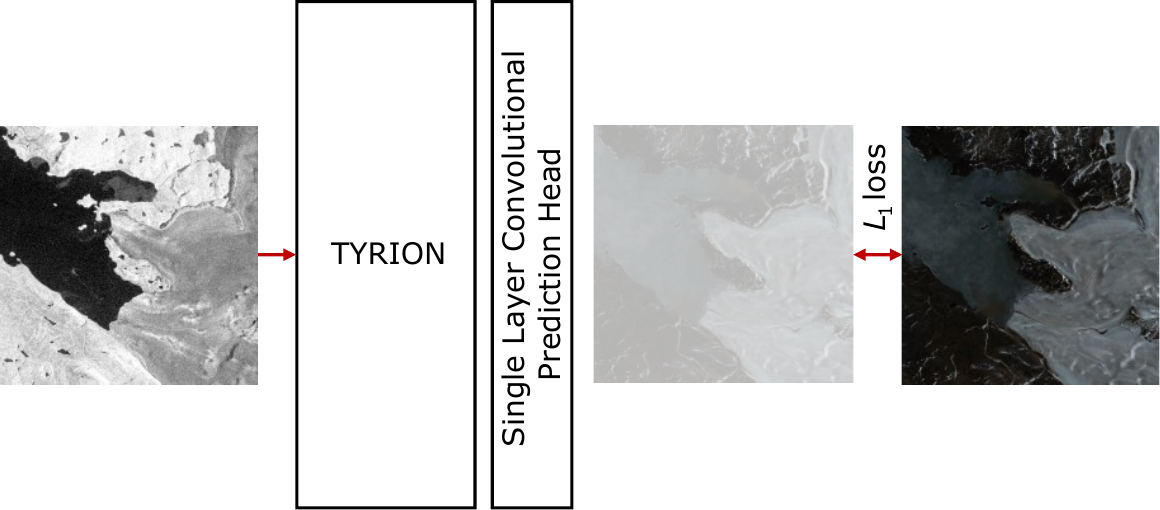}
            \caption{Visualization of SSL4SAR\,--\,OptTranslator. For illustration purposes, we show only the rgb channels of the optical image. The \ac{sar} image is shown without contrast enhancement to reflect the raw input provided to the model.}
            \label{fig:Translator}
        \end{figure}

        The second pretraining approach, is a straightforward generative \ac{ssl} method that employs TYRION with an additional single-layer convolutional head to translate each \ac{sar} image into its corresponding optical counterpart. SSL4SAR\,--\,OptTranslator is illustrated in Fig.~\ref{fig:Translator}.

\section{Experiments}
\label{sec:experiments}
To demonstrate the effectiveness of the two pretraining techniques and the TYRION architecture, we perform a series of experiments with and without pretraining and compare the results with those of the HookFormer~\cite{Wu.2023_2}, the current state-of-the-art model~\cite{Gourmelon.2025} for extracting calving fronts from \ac{sar}.

    \subsection{Evaluation Metrics}
    The primary evaluation metric for calving front delineation is the \ac{mde}~\cite{Gourmelon-2022}, which gives the symmetric mean distance between the predicted and ground truth calving front.
    The \ac{mde}~\cite{Gourmelon-2022} is calculated as:
    \begin{multline}\label{eq:mean_distance_error}
        \mathrm{MDE}(\mathcal{I}) = \frac{1}{\sum_{(\mathcal{P}, \mathcal{Q}) \in \mathcal{I}} (|\mathcal{P}| + |\mathcal{Q}|)}\\
        \cdot \sum_{(\mathcal{P}, \ \mathcal{Q}) \in I} \bigg( \sum_{\vec{p} \in \mathcal{P}} \min_{\vec{q} \in \mathcal{Q}} \lVert \vec{p}-\vec{q} \rVert_2 + \sum_{\vec{q} \in \mathcal{Q}} \min_{\vec{p} \in \mathcal{P}} \lVert \vec{p}-\vec{q} \rVert_2 \bigg)
    \end{multline}
    where $\mathcal{P}$ is the set of ground truth calving front pixels of an image, $\mathcal{Q}$ is the set of predicted calving front pixels of that image, and $\mathcal{I}$ is the set of all images. The cardinality of a set is represented as $|.|$ in the equation.
    
    \subsection{Training Setup}
\begin{figure}[ht]
        \centering
        \includegraphics[width=0.3\textwidth]{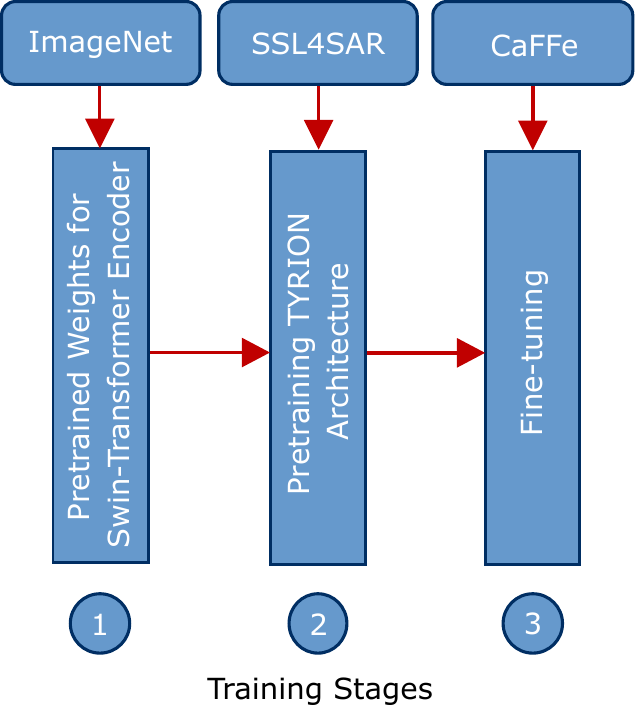}
        \caption{Visualization of the training pipeline. In stage 1, TYRION's Swin-Transformer Encoder is initialized with ImageNet pretrained weights. Stage 2 involves pretraining TYRION on SSL4SAR. In stage 3, TYRION is trained to segment landscape zones based on \ac{caffe}'s zone labels.}
        \label{fig:Training_Pipeline}
    \end{figure}

    A visualization of the training pipeline with the employed datasets can be found in Fig.~\ref{fig:Training_Pipeline}.
    The pipeline consists of three stages:  
    In stage 1, the model encoder is initialized with ImageNet pretrained weights obtained through supervised learning; if stage 1 is excluded, the encoder is trained from scratch.
    
    Stage 2 involves pretraining TYRION on SSL4SAR using either SSL4SAR\,--\,OptSimMIM or SSL4SAR\,--\,OptTranslator. 
    For pretraining (stage 2), the SSL4SAR dataset is split into a training set and a validation set. The latter comprises the initial 68 images in the time series of each of the 14 glaciers, representing approximately 10\,\% of the total data set.
    Random resized crops and flipping are applied on the fly to augment the images.
    An $L_1$ loss function is applied.
    
    In stage 3, TYRION is fine-tuned to segment landscape zones based on \ac{caffe}'s zone labels using the train-test split defined by Gourmelon et\,al.~\cite{Gourmelon-2022}.
    The weights for fine-tuning are initialized with the pretraining checkpoint from the epoch with the lowest $L_1$ loss on the validation set.
    The applied loss function is a combination of the cross-entropy on the full model output, the cross-entropy on the inner $256 \times 256$ of the output, the dice score on the full model output, and the dice score on the inner $256 \times 256$ of the output.
    We apply label smoothing ($\epsilon = 0.1$) for the cross-entropy losses.
    Augmentations for fine-tuning consist of random resized crops, vertical and horizontal flips, rotations, brightness adjustments, gamma corrections, additive Poisson noise, and a modified mixup variant that only alters the ocean area.
    For hyperparameter estimation, \ac{caffe}'s train set is further split into a 10\,\% validation set and a 90\,\% training set, while \ac{caffe}'s test set remains untouched until the final evaluation.
    For testing, the checkpoint with the lowest \ac{mde} on the validation set is chosen.
    Ultimately, the calving fronts are extracted from the segmented zones via the post-processing pipeline outlined by Gourmelon et\,al.~\cite{Gourmelon-2022}. This pipeline involves several steps: First, patch merging and the argmax operation are applied. Next, a connected components analysis is performed on the predicted ocean zone to retain only the largest ocean region and fill in any potential gaps or islands. The boundaries between this ocean region and predicted glacier regions are then extracted as potential calving fronts. Additionally, bounding box masking is applied to focus on the region of interest, and any calving fronts shorter than \SI{750}{\metre} are discarded.
    
    We conduct experiments using four different setups to evaluate the performance of our pretraining strategy and the TYRION architecture, and to analyze the overall impact and importance of pretraining.
        \begin{itemize}
            \item Setup 1 omits both Stage 1 and Stage 2, serving as a baseline.
            \item Setup 2 excludes only Stage 2 to assess the effect of domain-specific pretraining.
            \item Setup 3 omits Stage 1 to investigate the role of weight initialization.
            \item Setup 4 includes all three stages.
        \end{itemize}
    As a comparison to the TYRION architecture, the HookFormer~\cite{Wu.2023_2} is tested in two configurations: one corresponding to Setup 1 and the other to Setup 2.

    \subsection{Ensemble}
    In order to achieve the optimal results during the inference stage, it is essential to apply the model in the most effective manner.
    Therefore, in addition to the experiments demonstrating the superiority of the presented novel architecture and pretraining approaches, we apply well-established test-time techniques to further improve the results.
    These techniques include a sliding window approach with an overlap (50\,\%) for extracting patches from the test images, four-fold test-time augmentation through rotations, and the formation of an ensemble of five models by averaging their confidence values per image.
    The five models are re-trained versions of TYRION that have gone through the entire training pipeline, starting with the ImageNet pretrained weights (setup 2 or 4).
    Furthermore, the ensemble enables the calculation of uncertainty maps, which are generated by computing the pixel-wise standard deviation over the confidence values of the five models per class.
    
    \subsection{Multi-Annotator Evaluation}
    The \acp{mde} on the test set are calculated by comparing the results with \ac{caffe}'s front labels for all experiments.
    However, Gourmelon et\,al.~\cite{Gourmelon.2025} have conducted a multi-annotator study on \ac{caffe}'s test set, resulting in new averaged ground truth front labels and an inter-annotator \ac{mde}.
    In order to assess how close our models (setups 2 and 4) are to human performance, we apply the additional post-processing necessary for a fair comparison between humans and AI models, outlined by Gourmelon et\,al.~\cite{Gourmelon.2025}, and compare the predictions with the new averaged front labels.

\section{Results}
\label{sec:results}
\begin{table*}[htbp]
    \centering
    \caption{\acp{mde} in meters for \ac{caffe}'s test set divided by capturing sensor. A \textbf{bold font} signifies the best value within each division of a column. \emph{Weights} are the weights at initialization of the model. \textbf{Setup} refers to the configuration of the training pipeline. \emph{PALSAR} is short for ALOS Phased Array L-band Synthetic Aperture Radar. \emph{TSX} stands for TerraSAR-X and TanDEM-X}
    \label{tab:distance_errors_sensors_mde}
    \begin{tabular}{p{0.01\textwidth} p{0.085\textwidth} p{0.085\textwidth} p{0.06\textwidth} p{0.035\textwidth} p{0.075\textwidth} p{0.075\textwidth} p{0.075\textwidth} p{0.075\textwidth} p{0.075\textwidth} p{0.075\textwidth}} 
        \toprule
        &&&&& \emph{All} & \emph{Sentinel-1} & \emph{ENVISAT} & \emph{ERS} & \emph{PALSAR} & \emph{TSX} \\
        &\emph{Architecture} &\emph{pretraining} &\emph{Weights} &\emph{Setup} & & $\downarrow$ \emph{MDE} & $\downarrow$ \emph{MDE} & $\downarrow$ \emph{MDE} & $\downarrow$ \emph{MDE} & $\downarrow$ \emph{MDE} \\
        \midrule
        &\multirow{2}{*}{HookFormer} & / & Scratch & 1 & $938 \pm 90$ & $2654 \pm 277$ & $1004 \pm 364$ & $983 \pm 610$ & $678 \pm 208$ & $760 \pm 64$\\
        & & / & ImageNet & 2 & $360 \pm 13$ & $918 \pm 76$ & $253 \pm 42$ & $174 \pm 47$ & $263 \pm 28$ & $286 \pm 8$ \\
        \midrule
        &\multirow{6}{*}{TYRION}
        & / & Scratch & 1 & $682 \pm 101$ & $1604 \pm 230$ & $644 \pm 148$ & $512 \pm 520$ & $346 \pm 114$ & $550 \pm 97$ \\
        && / & ImageNet & 2 & $342 \pm 25$ & $750 \pm 79$ & $325 \pm 96$ & $168 \pm 50$ & $184 \pm 45$ & $274 \pm 24$ \\
        &&\multirow{ 2}{*}{OptSimMIM} & Scratch & 3 & $773 \pm 92$ & $1961 \pm 407$ & $598 \pm 264$ & $448 \pm 357$ & $364 \pm 161$ & $610 \pm 76$ \\
        &&& ImageNet & 4 & $300 \pm 60$ & $\mathbf{592 \pm 175}$ & $307 \pm 123$ & $154 \pm 68$ & $192 \pm 65$ & $253 \pm 42$ \\
        &&\multirow{ 2}{*}{OptTranslator} & Scratch & 3 & $574 \pm 92$ & $1364 \pm 175$ & $629 \pm 148$ & $485 \pm 358$ & $338 \pm 109$ & $457 \pm 90$ \\
        &&& ImageNet & 4 & $\mathbf{293 \pm 54}$ & $638 \pm 309$ & $\mathbf{234 \pm 67}$ & $\mathbf{148 \pm 68}$ & $\mathbf{182 \pm 78}$ & $\mathbf{240 \pm 14}$ \\
        \midrule
        \multirow{3}{*}{\rotatebox{90}{\emph{Ensem.}}}
        &\multirow{3}{*}{TYRION} &/& ImageNet & 2 & $308$ & $674$ & $587$ & $137$ & $200$ & $243$ \\
        &&OptSimMIM & ImageNet & 4 & $\mathbf{238}$ & $\mathbf{398}$ & $456$ & $\mathbf{70}$ & $\mathbf{142}$ & $\mathbf{209}$ \\
        &&OptTranslator & ImageNet & 4 & $246$ & $425$ & $\mathbf{296}$ & $\mathbf{70}$ & $147$ & $217$ \\
        \midrule
        \multicolumn{11}{c}{\emph{Multi-Annotator Study}}\\
        &HookFormer & / & ImageNet & 2 & $221 \pm 15$ & $915 \pm 108$ & $230 \pm 40$ & $157 \pm 46$ & $186 \pm 32$ & $105 \pm 5$\\
        \multirow{3}{*}{\rotatebox{90}{\emph{Ensem.}}} &\multirow{3}{*}{TYRION} &/& ImageNet & 2 & $131$ & $437$ & $591$ & $108$ & $161$ & $67$ \\
        &&OptSimMIM & ImageNet & 4 & $86$ & $241$ & $476$ & $47$ & $104$ & $49$ \\
        &&OptTranslator & ImageNet & 4 & $75$ & $254$ & $285$ & $53$ & $\mathbf{100}$ & $38$ \\
        & Humans & & & & $\mathbf{38 \pm 15}$ & $\mathbf{99 \pm 26}$ & $\mathbf{151 \pm 228}$ & $\mathbf{22 \pm 6}$ & $129 \pm 182$ & $\mathbf{21 \pm 6}$ \\
        \bottomrule
        &&&& & & & & & & \\
    \end{tabular}
\end{table*}

\begin{figure*}[htbp]
    \centering
    \includegraphics[width=\textwidth]{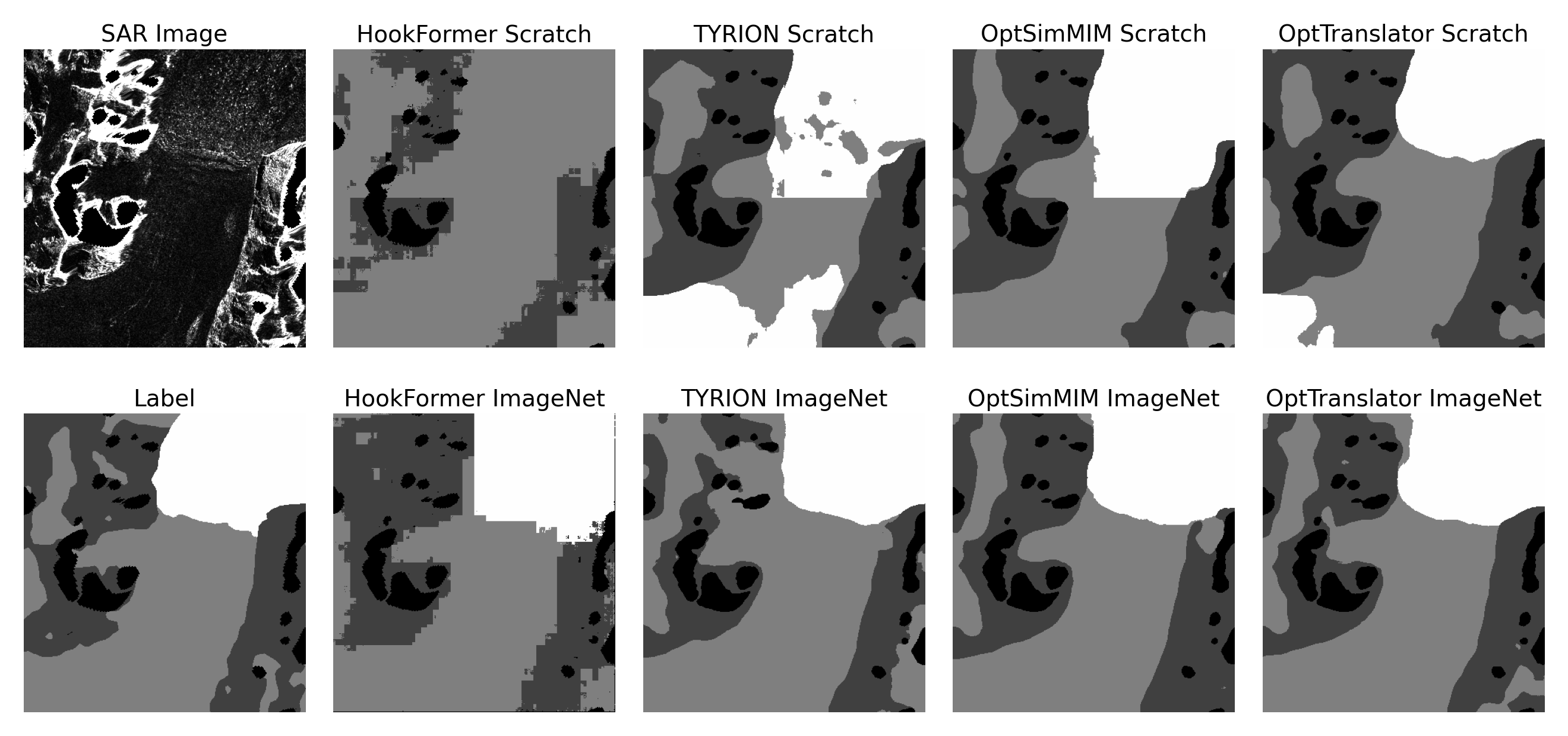}
    \caption{Sample image of the Mapple Glacier captured by \ac{s1} on January 8, 2020 with the corresponding zone label and zone segmentations of the experiments. The ocean is visualized in white, the glacier in light gray, rock in dark gray, and black indicates areas where no information is available. The \ac{sar} image is shown without contrast enhancement to reflect the raw input provided to the model.}
    \label{fig:zones}
\end{figure*}

\begin{figure*}[htbp]
    \centering
    \includegraphics[width=\textwidth]{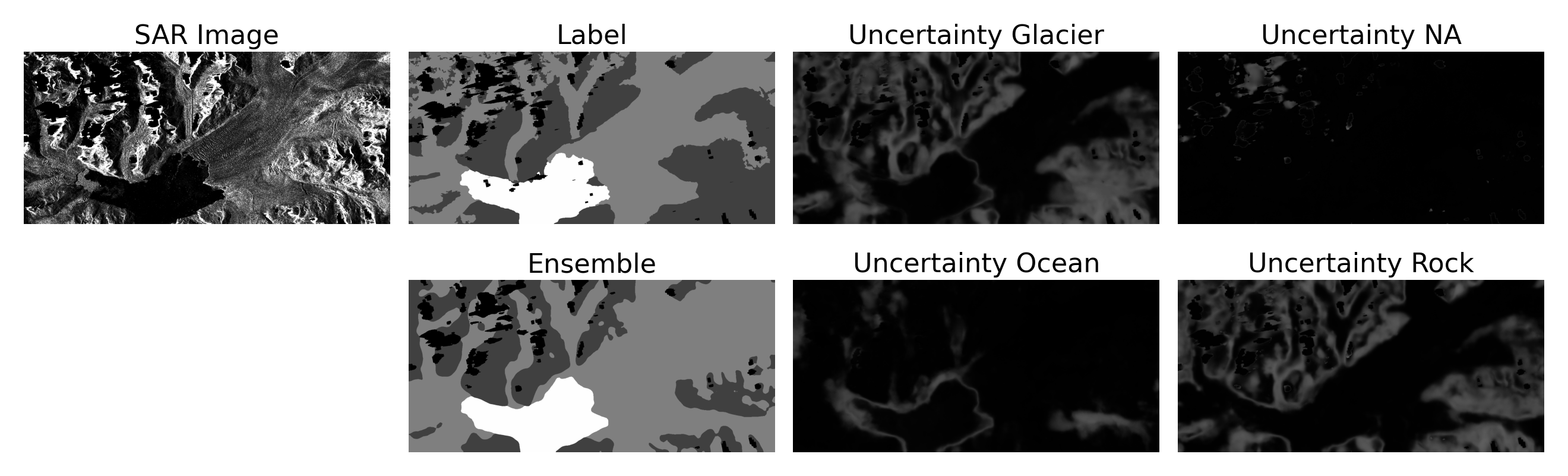}
    \caption{Sample image of the Columbia Glacier captured by \ac{s1} on August 14, 2016 with the corresponding zone label, and the ensemble TYRION OptTranslator (setup 4) model's zone segmentation with uncertainty maps per class. The \ac{sar} image is shown without contrast enhancement to reflect the raw input provided to the model.}
    \label{fig:uncertainty}
\end{figure*}

\begin{figure}[htbp]
    \centering
    \begin{tikzpicture}[scale=1.0, declare function={
            colorcalc(\nofront) = \nofront;
      }],
        \centering
        \begin{axis}[
        legend style={
              at={(1,0)},yshift=165pt,xshift=1em,anchor=north east,
              my area legend,
              name=legend,
              draw={none}, 
              fill={none},
            },
        legend image code/.code={%
            \draw[#1, draw=none] (0cm,-0.08cm) rectangle (0.4cm,0.08cm);
        },  
        height=8cm,
        width=0.48\textwidth,  
        bar width=20,
        ymax=280,
        ymin=0,
        xmin=0.5,
        xmax=8.5,
        ymajorgrids=true,
        axis y line*=left,
        axis x line*=bottom,
        xticklabels={HookFormer, TYRION, OptSimMIM, OptTranslator, Ensem. TYRION, Ensem. OptSimMIM, Ensem. OptTranslator, Humans},
        xtick={1, 2, 3, 4, 5, 6, 7, 8},
        x tick label style={rotate=50,anchor=east, font=\small},
        y tick label style={font=\small},
        ylabel={\small Mean Distance Error (m)}]
            \foreach \x/\mean/\nofront [evaluate=\nofront as \color using {colorcalc(\nofront)}] in {1/221/100, 2/168/100, 3/151/100, 4/124/100, 5/131/100, 6/86/100, 7/75/100, 8/38/100} {
                \edef\temp{\noexpand\addplot[ybar,coolblack,forget plot,fill=bluegray!\color] coordinates {(\x,\mean)};
                }
                \temp
            }
            
            \addplot+[only marks,black,mark options={draw=black,fill=black},forget plot][error bars/.cd,y dir=both, y explicit]
            coordinates {
            
            (1,221) +- (0,21)
            (2,168) +- (0,39)
            (3,151) +- (0,123)
            (4,124) +- (0,73)
            (8,38) +- (0,12)
            };            
        \end{axis}
    \end{tikzpicture}%
    \caption{Overview of the \acp{mde} for the multi-annotator study and the results of the post-processed HookFormer (setup 2), TYRION (setup 2) and TYRION with the two pretraining approaches -- OptSimMIM and OptTranslator -- (setup 4), as well as the ensembles of the latter three. Confidence intervals are provided for the \acp{mde} of humans and individual models (excluding the ensembles).}
    \label{fig:human_vs_ai}
\end{figure}
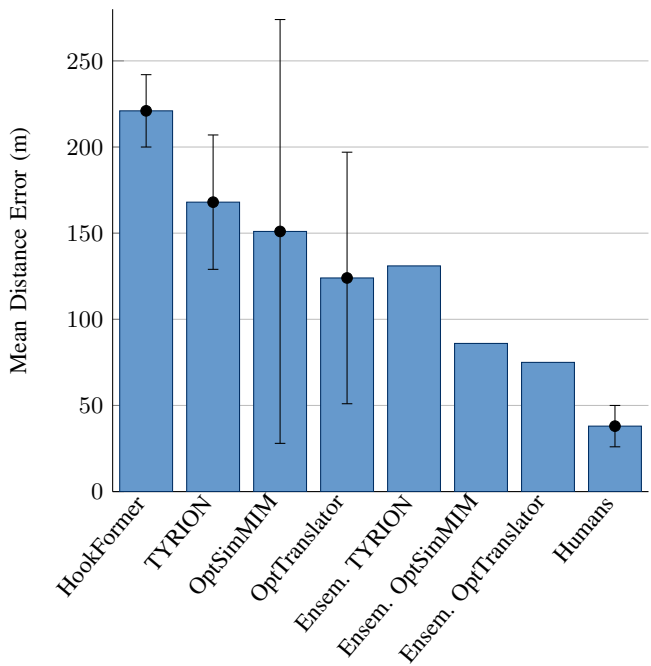

\begin{figure*}[htbp]
    \centering
    \includegraphics[width=\textwidth]{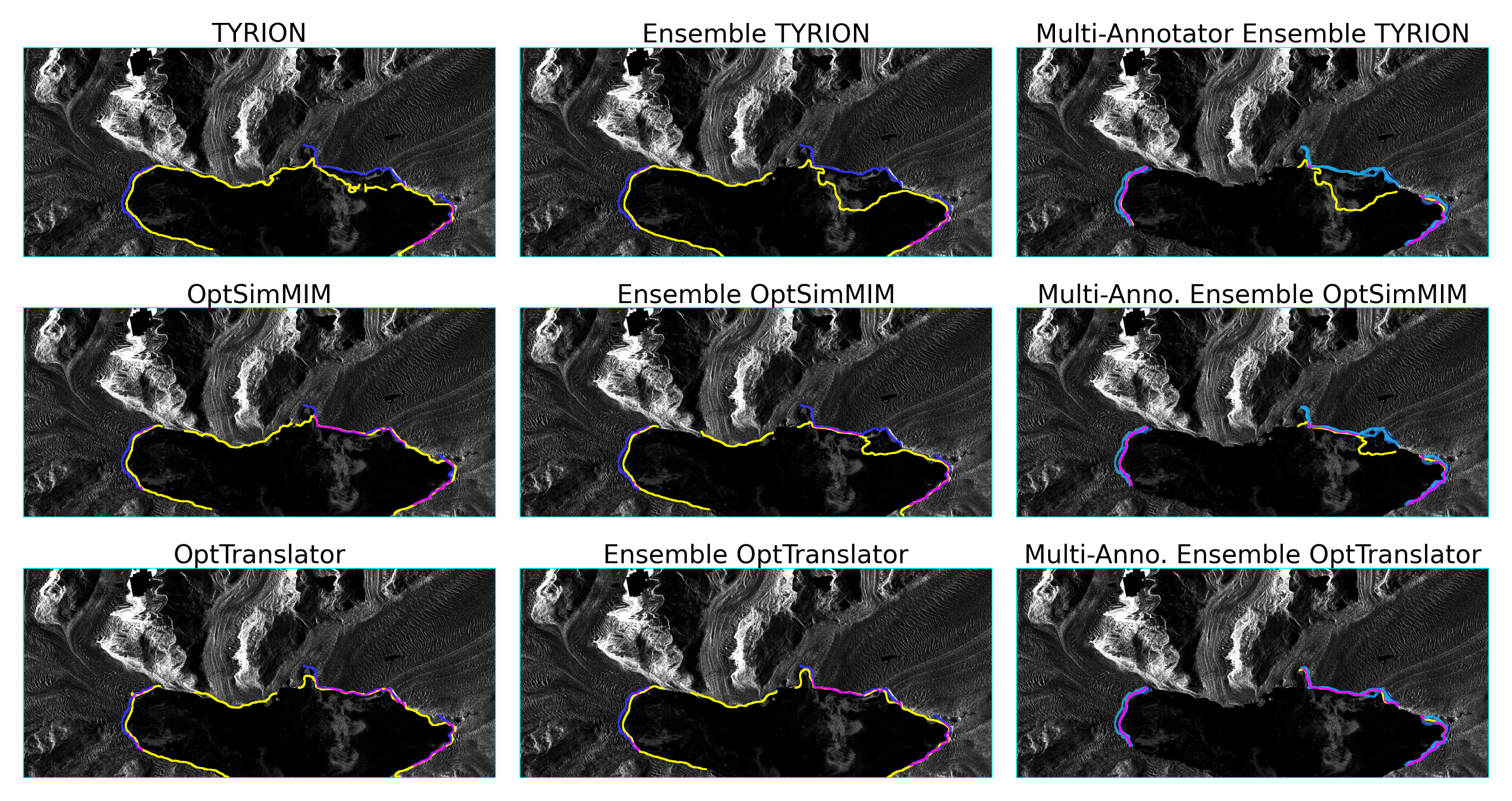}
    \caption{Visualizations of a sample image of Columbia Glacier taken by \ac{tdx} on October 15, 2014 for TYRION (setup 2) and TYRION with the two pretraining approaches -- OptSimMIM and OptTranslator -- (setup 4), their corresponding ensembles, and post-processed ensembles for the multi-annotator study. \fboxsep=1pt\colorbox{yellow!100}{Yellow} represents the prediction, \fboxsep=1pt\colorbox{blue!100}{\color{white}blue} is used for the ground truth front, different \fboxsep=1pt\colorbox{bleudefrance!100}{\color{white}shades of blue} for all ten annotators of the multi-annotator study, and \fboxsep=1pt\colorbox{magenta!100}{\color{white}pink} represents a perfect match between prediction and ground truth/annotator. The bounding box is given by \fboxsep=1pt\colorbox{Turquoise!100}{turquoise}. The \ac{sar} image is shown without contrast enhancement to reflect the raw input provided to the model.}
    \label{fig:all}
\end{figure*}

The quantitative results on the test set of the \ac{caffe} dataset for all experiments are summarized in Table~\ref{tab:distance_errors_sensors_mde}, which also provides a sensor-specific breakdown of the results.
The lowest average \ac{mde} of \SI{293}{\metre} is achieved by TYRION SSL4SAR\,--\,OptTranslator initialized with ImageNet-pretrained weights (setup 4).
Qualitative results are visualized in Fig.~\ref{fig:zones}, which displays a sample image with its zone label and zone segmentations of the experiments, and Fig.~\ref{fig:all}, illustrating the predicted calving fronts for one sample image across TYRION (setup 2) and the two pretraining approaches -- OptSimMIM and OptTranslator -- (setup 4), along with their ensemble and post-processed ensemble results for the multi-annotator study.

    \subsection{Architecture comparison: TYRION vs. HookFormer}
    Even without pretraining, TYRION (setup 2) outperforms the previous state of the art, the HookFormer~\cite{Wu.2023_2} (setup 2), with a mean \ac{mde} improvement of \SI{18}{\metre}. 
    As intended with the \ac{CNN} decoder, TYRION produces smoother calving front predictions compared to the HookFormer. 
    As shown in Figure~\ref{fig:zones}, the HookFormer prediction exhibits jagged edges at the interface between the ocean and glacier, whereas the TYRION output exhibits considerably smoother delineations.
    Additionally, the number of parameters has been reduced; while the HookFormer has 59.3\,G parameters, the TYRION architecture has only 50.9\,G parameters.

    \subsection{Impact of pretraining}
    The integration of pretraining consistently improves TYRION's performance, irrespective of the \ac{ssl} approach employed (setup 2 versus 4).
    For SSL4SAR\,--\,OptSimMIM, the \ac{mde} decreases by \SI{42}{\metre}, while SSL4SAR\,--\,OptTranslator achieves a reduction of \SI{49}{\metre}.
    
    Qualitative improvements can be seen in figures~\ref{fig:zones} and~\ref{fig:all}.
    When comparing the pretrained TYRION models (setup 4) to HookFormer (setup 2), the \ac{mde} differences are larger, with \SI{60}{\metre} and \SI{67}{\metre} reductions for SSL4SAR\,--\,OptSimMIM and SSL4SAR\,--\,OptTranslator, respectively. 
    The observed difference between the two pretraining approaches is low with \SI{7}{\metre}.

    \subsection{Role of Initialization: ImageNet-pretrained Weights}
    Training the HookFormer from scratch (setup 1 versus 2) more than doubles its \ac{mde} and leads to an average of 25 images where no front is recognized at all out of the 122 images in the test set.
    Initialization with ImageNet-pretrained weights also enhances TYRION's performance (setup 3 versus 4 and setup 1 versus 2), both with and without the introduced \ac{ssl} pretraining approaches. TYRION initialized with ImageNet-pretrained weights (setup 2) outperforms both TYRION trained from scratch (setup 1) and TYRION pretrained on SSL4SAR starting from scratch (setup 3). 
    The benefits of initialization are evident in Figure~\ref{fig:zones}, where predictions initialized with ImageNet weights demonstrate a notable improvement in discerning between glacier and ocean regions.

    \subsection{Sensor-Specific Analysis}
    \label{subsec:sensor_analysis}
    The impact of pretraining varies by sensor, as detailed in Table~\ref{tab:distance_errors_sensors_mde}. 
    Pretraining with SSL4SAR\,--\,OptSimMIM (setup 2 versus 4) achieves the greatest \ac{mde} reduction for \ac{s1} images (21\,\%), whereas changes for other sensors range between a reduction of 8\,\% and a slight increase of 4\,\%. In contrast, SSL4SAR\,--\,OptTranslator (setup 2 versus 4) yields the largest reduction for ENVISAT images (28\,\%), followed by \ac{s1} images (15\,\%). Overall, pretraining on SSL4SAR provides the greatest performance gains for \ac{s1} images.

    \subsection{Ensemble Approach}
    The ensemble approach, which leverages test-time augmentations and overlapping patches, yields further reductions in \ac{mde}. For TYRION without pretraining (setup 2), the ensemble approach decreases the \ac{mde} by \SI{34}{\metre}, while reductions of \SI{62}{\metre} and \SI{47}{\metre} are observed for SSL4SAR\,--\,OptSimMIM and SSL4SAR\,--\,OptTranslator (setup 4), respectively.  
    Sensor-specific analysis reveals that ensemble approaches reduce the \ac{mde} for most sensors, except ENVISAT, where all three configurations -- TYRION without pretraining and with the two pretraining approaches (setups 2 and 4) -- show an increase in \ac{mde}. 
    For PALSAR, only TYRION without pretraining (setup 2) results in an increase in \ac{mde}. 
    Figure~\ref{fig:uncertainty} displays the uncertainty maps generated using the ensemble approach. 
    Of particular interest is the uncertainty map for the ocean class, which effectively illustrates the model's confidence in delineating the ocean boundaries, with the exception of the left calving front, where there is a discernible degree of uncertainty regarding the precise location of the ocean edge. 
    The importance of each component of the ensemble approach is demonstrated in the experiments presented in Sec.~\ref{sec:ablation}.

    \subsection{Multi-Annotator Study}
    Finally, Table~\ref{tab:distance_errors_sensors_mde} presents the results of the multi-annotator study. 
    The TYRION ensemble model pretrained with SSL4SAR\,--\,OptTranslator (setup 4) achieves an \ac{mde} of \SI{75}{\metre} on the entire test set, which is only \SI{37}{\metre} more than the inter-human \ac{mde} of \SI{38}{\metre}. 
    Predictions for ENVISAT images exhibit the highest \acp{mde}, with a mean deviation from the averaged human annotators' \ac{mde} of \SI{325}{\metre} and \SI{134}{\metre}, depending on the pretraining method applied (setup 4).
    In contrast, TYRION pretrained with SSL4SAR\,--\,OptTranslator (setup 4) even surpasses human performance on PALSAR images, achieving an \ac{mde} of \SI{100}{\metre} compared to \SI{129}{\metre} for humans. 
    The results of the multi-annotator study are represented graphically in Fig.~\ref{fig:human_vs_ai}, showing the performance improvements achieved by the novel architecture, the multimodal pretraining techniques, and the ensemble approach.
    The right side of Fig.~\ref{fig:all} illustrates how close the model predictions are to the human annotations, while minor misinterpretations still leave room for improvement.

\subsection{Ablation Studies}
\label{sec:ablation}
We perform an ablation study on the components of our ensemble approach (setup 4), i.\,e., we remove the test-time augmentation, additional overlap, and the ensembling itself one by one and check the influence on the performance metrics.
The results are given in Tab.~\ref{tab:statistics}.
While \acp{mde} decrease with the additional techniques, the throughput time per image increases.
When using ensembling, the model size increases five-fold.
In general, test-time augmentations appear to offer greater performance improvements than overlapping patches. 
For SSL4SAR\,--\,OptTranslator, the impact of test-time augmentations even surpasses the performance gains achieved through ensembling. 
Overlapping patches provide a substantial performance boost for SSL4SAR\,--\,OptTranslator only in the absence of test-time augmentations; otherwise, their contribution is more limited.
In the case of SSL4SAR\,--\,SimMIM, ensembling yields a greater performance improvement compared to SSL4SAR\,--\,OptTranslator.
However, overlapping patches and test-time augmentations are less effective for SSL4SAR\,--\,SimMIM than for SSL4SAR\,--\,OptTranslator. 
Nevertheless, the overall performance gain achieved using the complete ensemble approach is higher for SSL4SAR\,--\,SimMIM than for SSL4SAR\,--\,OptTranslator.

\begin{table*}[htbp]
    \newcolumntype{P}[1]{>{\centering\arraybackslash}p{#1}}
    \centering
    \caption{Ablation study on the components of the ensemble approach (setup 4). The throughput denotes the time of the model including post-processing for the entire test set divided by the number of images in the test set. All models were initialized with ImageNet weights. The \ac{mde} is given in meters. A \textbf{bold font} indicates the best value within a column.}
    \label{tab:statistics}
    \begin{tabular}{p{0.07\textwidth} p{0.09\textwidth} P{0.09\textwidth} P{0.09\textwidth} P{0.09\textwidth} P{0.2\textwidth} P{0.1\textwidth} } 
        \toprule
        \emph{Model} & \emph{Pretraining} & \emph{Ensemble} & \emph{Overlap} & \emph{Test-time Aug.} & $\uparrow$
        \emph{Throughput [image/min]} & $\downarrow$ \emph{MDE}\\ 
        \midrule
        \multirow{10}{*}{TYRION} & OptSimMIM&&&& $11.55$ & $300$\\ 
        & OptTranslator&&&& $\mathbf{11.61}$ & $293$\\ 
        & OptSimMIM & \checkmark &&& $\phantom{0}1.00$ & $263$ \\  
        & OptTranslator & \checkmark &&& $\phantom{0}0.94$ & $278$ \\ 
        & OptSimMIM & \checkmark & \checkmark && $\phantom{0}0.49$ & $259$ \\ 
        & OptTranslator & \checkmark & \checkmark && $\phantom{0}0.50$ & $262$\\ 
        & OptSimMIM & \checkmark && \checkmark & $\phantom{0}0.24$ & $242$\\ 
        & OptTranslator & \checkmark && \checkmark & $\phantom{0}0.23$ & $248$\\ 
        & OptSimMIM & \checkmark & \checkmark & \checkmark & $\phantom{0}0.13$ & $\mathbf{238}$\\ 
        & OptTranslator & \checkmark & \checkmark & \checkmark & $\phantom{0}0.13$ & $246$\\ 
        \bottomrule
    \end{tabular}
\end{table*}

\subsection{Discussion}
\label{sec:discussion}
In terms of pretraining, initializing TYRION with ImageNet-pretrained weights without subsequent pretraining on SSL4SAR (setup 2) outperforms \ac{ssl} pretraining from scratch (setup 3).
On the other hand, \ac{ssl} pretraining starting from ImageNet-pretrained weights (setup 4) outperforms just using ImageNet-pretrained weights (setup 2).
This leads us to conclude that supervised pretraining on ImageNet yields more valuable information than \ac{ssl} pretraining on SSL4SAR. 
There are two possible reasons for this: 
\begin{enumerate}
    \item the sheer size of ImageNet -- over 14\,million images versus SSL4SAR's 9,562 images, and
    \item the difference between supervised and \acf{ssl}.
\end{enumerate}
Nevertheless, subsequent pretraining of models initialized on ImageNet-pretrained weights with data from the target domain, which are provided by SSL4SAR, still improves the ImageNet-pretrained weights, indicating that some information in SSL4SAR is not present in ImageNet.
This is in line with previous works that stress the importance of in-domain pretraining data~\cite{Ren.2024, Dimitrovski.2024, Wang.2022, Manas.2021, Li.2023}.

The substantial decline in \ac{mde} for \ac{s1} images resulting from pretraining can be attributed to the fact that SSL4SAR encompasses solely images from \ac{s1}. 
If images from other sensors had been incorporated into the dataset, the discrepancy in performance increase between sensors might have been less pronounced.
Nevertheless, other sensors also benefit considerably from pretraining on SSL4SAR.

Taking the ensemble leads to an increase in \ac{mde} for images captured by ENVISAT.
Moreover, ENVISAT images have the highest \ac{mde} in the multi-annotator study, surpassing all other sensors.
One potential explanation for this phenomenon is that the images of this sensor also pose the most difficulties and ambiguities for human annotators, resulting in the high inter-human \ac{mde} of \SI{151}{\metre} with a substantial standard deviation of \SI{228}{\metre}.

\section{Conclusion}
\label{sec:conclusion}
This study presents a novel, state-of-the-art deep learning model for delineating glacier calving fronts in \ac{sar} imagery.
The model is based on an optimized architecture mixing a Swin-Transformer and a residual \ac{CNN} decoder.
An enhanced ensemble approach further improves the results and enables the generation of uncertainty maps.
Furthermore, the study proposes two novel multimodal, \ac{ssl}
pretraining strategies for \ac{sar} input data that require significantly less optical data than other multimodal remote sensing pretraining approaches.
To facilitate the pretraining strategies, we introduce a new dataset, SSL4SAR, containing 9,562 \ac{s1} images of 14 Arctic glaciers and one additional \ac{s2} image per glacier.
By initializing the ensemble model with ImageNet-pretrained weights and subsequently pretraining it on SSL4SAR, the model achieves an \ac{mde} of \SI{238}{\metre} on the \ac{caffe} benchmark.
The results emphasize the importance of pretraining on an in-domain dataset in addition to supervised pretraining that leverages large, labeled datasets such as ImageNet.
With supplementary post-processing from the multi-annotator study, the model achieves an MDE of \SI{75}{\metre} when compared to the ground truth from the same study, coming close to the human performance of \SI{38}{\metre}.

\section*{Code and Data Availability}
The code is publicly available on GitHub at \url{https://github.com/Nora-Go/TYRION} after acceptance.
The SSL4SAR dataset is publicly available from zenodo at \url{https://doi.org/10.5281/zenodo.14748506}.
The CaFFe benchmark dataset is publicly available from PANGAEA at \url{https://doi.pangaea.de/10.1594/PANGAEA.940950}.

\section*{Acknowledgments}
This research was funded by the Bayerisches Staatsministerium für Wissenschaft und Kunst within the Elite Network Bavaria with the Int. Doct. Program ``Measuring and Modelling Mountain Glaciers in a Changing Climate'' (IDP M3OCCA) as well as the German Research Foundation (DFG) project ``Large-scale Automatic Calving Front Segmentation and Frontal Ablation Analysis of Arctic Glaciers using Synthetic-Aperture Radar Image Sequences (LASSI)'' and the project ``PAGE'' within the DFG Emmy-Noether-Programme.
The authors gratefully acknowledge the scientific support and HPC resources provided by the Erlangen National High Performance Computing Center (NHR@FAU) of the Friedrich-Alexander-Universität Erlangen-Nürnberg (FAU) under the NHR projects b110dc. NHR funding is provided by federal and Bavarian state authorities. NHR@FAU hardware is partially funded by the DFG – 440719683.
The author team acknowledges the provision of satellite data under various AOs from respective space agencies (DLR, ESA, JAXA, CSA).
During the preparation of this work, AI technologies were used to assist in the writing process. Specifically, Grammarly (Grammarly, Inc., San Francisco, CA, USA) was used in order to check for grammar and style consistency and DeepL (DeepL SE, Cologne, Germany) and ChatGPT (GPT‐4) (OpenAI, San Francisco, CA, USA) were used in order to assist with rephrasing and improving readability. After using these tools, the manuscript was carefully reviewed and the content was edited as needed. No tools or services were used for content generation.

\section*{Author contribution}
\noindent \textbf{Nora Gourmelon}: Conceptualization, Methodology, Software, Experiments, Statistical Analysis, Project administration, Writing - Original draft preparation.  \textbf{Marcel Dreier}: Methodology, Software, Writing - review \& editing. \textbf{Martin Mayr}: Validation, Writing - review \& editing. \textbf{Thorsten Seehaus}: Supervision, Data curation, Writing - review \& editing. \textbf{Dakota Pyles}: Writing - review \& editing. \textbf{Matthias Braun}: Supervision, Writing – review \& editing. \textbf{Andreas Maier}: Supervision, Writing – review \& editing. \textbf{Vincent Christlein}: Supervision, Writing - review \& editing.

\bibliographystyle{IEEEtran}
\bibliography{IEEEabrv,bib}

\newpage

\section{Biography Section}

\begin{IEEEbiography}[{\includegraphics[width=1in,height=1.25in,clip,keepaspectratio]{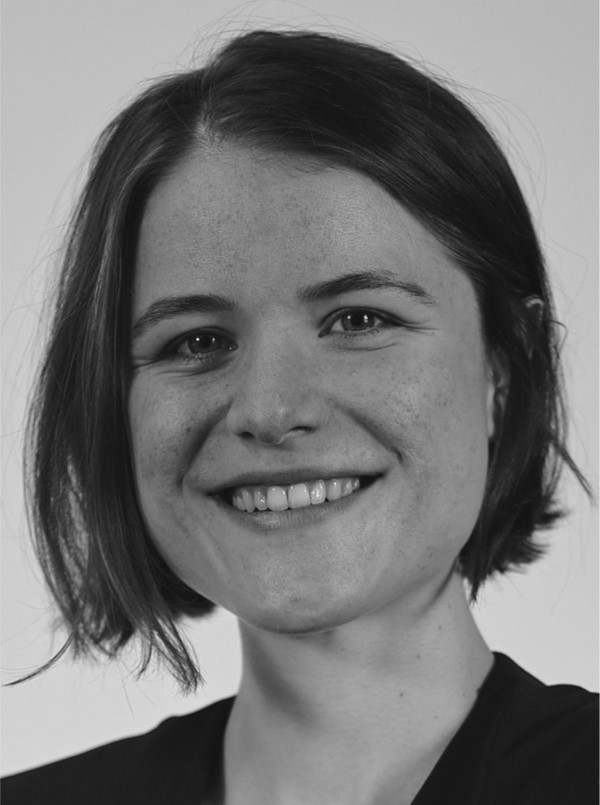}}]{Nora Gourmelon}
received the B.Sc. and M.Sc. degrees (passed with distinction) in computer science from Friedrich-Alexander-Universität Erlangen-Nürnberg (FAU), Erlangen, Germany, in 2019 and 2020, respectively, where she is currently pursuing the Ph.D. degree in computer science with the Pattern Recognition Laboratory (PRL).
She joined the PRL, FAU, in 2020, and is part of the International Doctorate Program ``Measuring and Modeling Mountain glaciers and ice caps in a Changing ClimAte'' (M3OCCA).
She was honored as AI Newcomer 2023 in the field of natural and life sciences by the German Association of Computer Science.
Her main research interests include applications of AI on topics related to sustainability and natural sciences.
\end{IEEEbiography}

\begin{IEEEbiography}[{\includegraphics[width=1in,height=1.25in,clip,keepaspectratio]{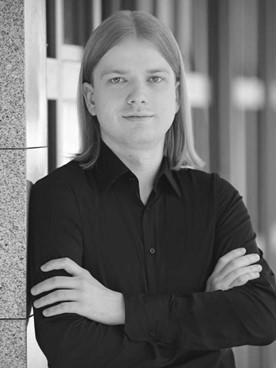}}]{Marcel Dreier}
earned his Bachelor’s and Master’s degree in computer science at FAU. In his Master's thesis he used diffusion models to generate offline handwritten text images and completed it in August 2023. Later on, he joined the Pattern Recognition Lab in October 2023 as a Ph.D candidate under the supervision of Prof. Andreas Maier. His current research focuses on machine learning on radargrams. 
\end{IEEEbiography}

\begin{IEEEbiography}[{\includegraphics[width=1in,height=1.25in,clip,keepaspectratio]{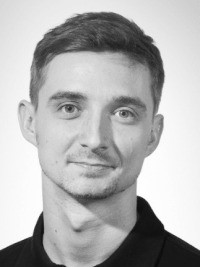}}]{Martin Mayr}
obtained his Master’s degree in Computer Science from Friedrich–Alexander University Erlangen–Nürnberg, Germany, in 2019. He is currently a Ph.D. candidate in the university’s Computer Vision Group, focusing on handwritten text recognition, handwriting imitation, and writer identification. In addition to his research, he works at the Erlangen National High Performance Computing Center, where he is involved in teaching and providing support in artificial intelligence.
\end{IEEEbiography}

\begin{IEEEbiography}[{\includegraphics[width=1in,height=1.25in,clip,keepaspectratio]{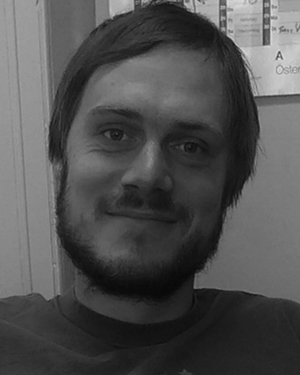}}]{Thorsten Seehaus}
received the Diploma degree in physics from the University of Würzburg, Würzburg, Germany, in 2011, and the Ph.D. degree in geography from FAU, Erlangen, Germany, in 2016.
He finished an apprenticeship as a Mechatronics Technician at Jopp GmbH, Bad Neustadt an der Saale, Germany, in 2003. In 2012, he joined the Working Group of Geographic Information System (GIS) and Remote Sensing, Institute of Geography, FAU, where he is currently a Junior Research Group Leader. He uses mainly multimission synthetic aperture radar (SAR) imagery to assess glacier variables, such as mass balances and area changes. His research interests include developing and applying remote sensing techniques for monitoring glacier changes on various scales and in various regions worldwide.
\end{IEEEbiography}

\begin{IEEEbiography}[{\includegraphics[width=1in,height=1.25in,clip,keepaspectratio]{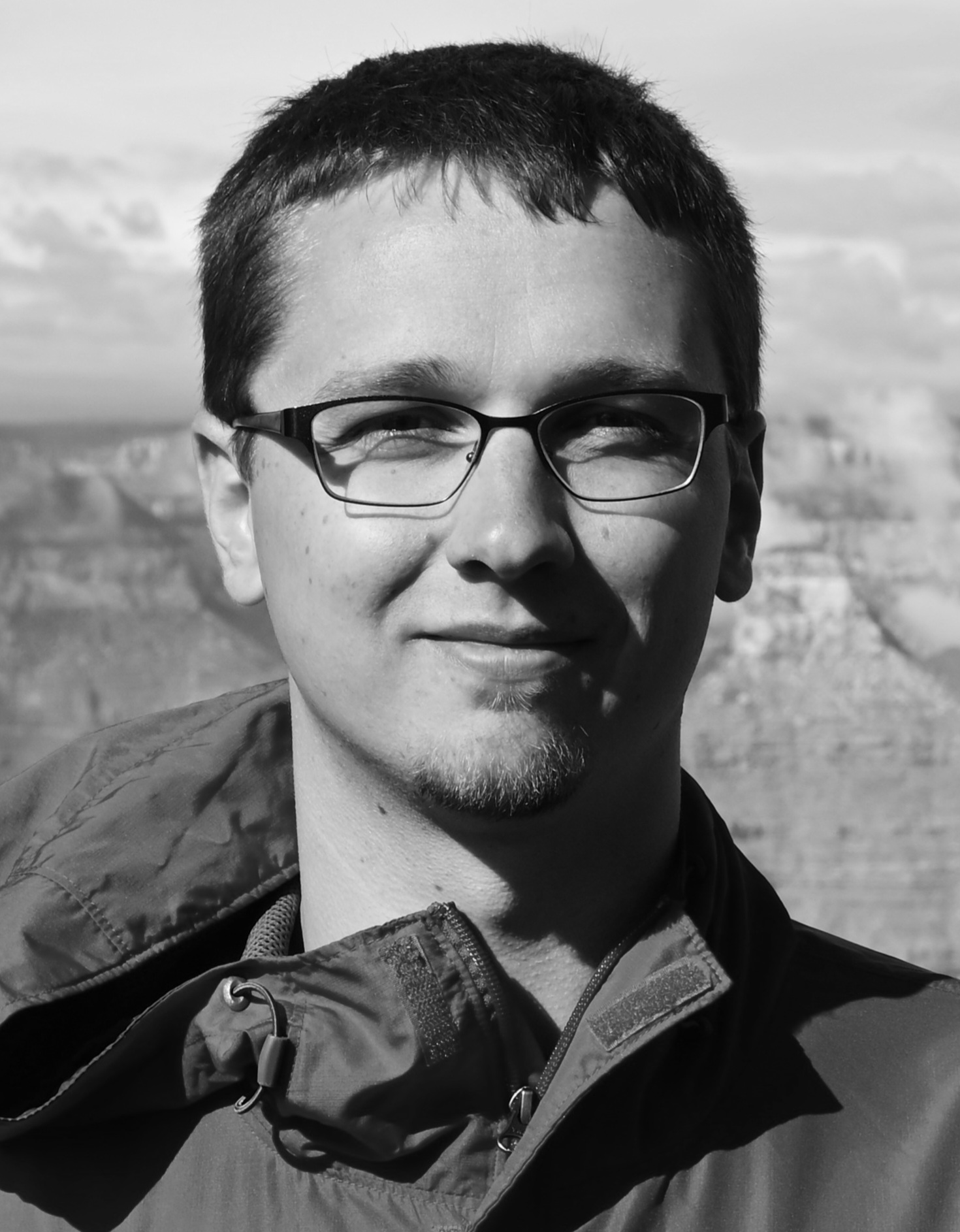}}]{Dakota Pyles}
received a B.Sc. in Geosciences from the University of Montana in 2019 and a M.Sc. in Geology from the University of Idaho in 2022. He is currently pursuing a Ph.D. degree in the Working Group of Geographic Information Systems (GIS) and Remote Sensing at the FAU. In 2023, he joined the Institute of Geography, FAU, and is affiliated with the International Doctorate Program M3OCCA. His current research focuses on estimating frontal ablation in the Arctic and understanding spatiotemporal drivers of observed tidewater glacier changes. 
\end{IEEEbiography}

\begin{IEEEbiography}[{\includegraphics[width=1in,height=1.25in,clip,keepaspectratio]{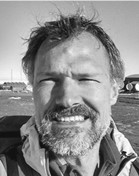}}]{Matthias Braun}
received the Diploma degree in hydrology and the Dr. rer.nat. (Ph.D.) degree (Hons.) from the University of Freiburg, Breisgau, Germany, in 1997 and 2001, respectively. From 2001 to 2010, he was the Scientific Coordinator of the interdisciplinary Center for Remote Sensing of Land Surfaces at Bonn University, Germany. He was appointed as an Associate Professor of geophysics with the University of Alaska Fairbanks, Fairbanks, AK, USA, in 2010, and as a Professor with FAU, Germany, in 2011. His research interests cover mass change of glaciers for which he combines in-situ observations, remote sensing and modelling with a strong focus on large-scale Earth observation data analysis. He has been leading numerous field campaigns in Antarctica, Greenland, Svalbard, Patagonia, High Mountain Asia, and the Alps. 
\end{IEEEbiography}

\begin{IEEEbiography}[{\includegraphics[width=1in,height=1.25in,clip,keepaspectratio]{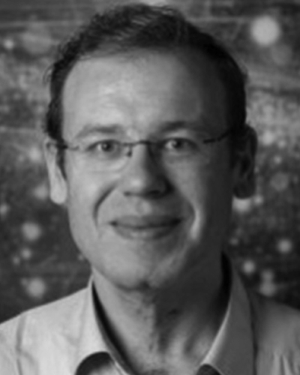}}]{Andreas Maier}
(Senior Member, IEEE) was born in Erlangen, Germany, in November 1980. He graduated in computer science and the Ph.D. degree from FAU, Erlangen, in 2005 and 2009, respectively.
From 2005 to 2009, he was with the PRL, Computer Science Department, FAU. His major research subject was medical signal processing in speech data. In this period, he developed the first online speech intelligibility assessment tool—PEAKS—that has been used to analyze over 4000 patients and control subjects so far. From 2009 to 2010, he started working on flat-panel C-arm CT as a Post-Doctoral Fellow at the Radiological Sciences Laboratory, Department of Radiology, Stanford University, Stanford, CA, USA. From 2011 to 2012, he was with Siemens Healthcare, Erlangen, Germany, as the Innovation Project Manager and was responsible for reconstruction topics in the angiography and X-ray business unit. In 2012, he returned to FAU as the Head of the Medical Reconstruction Group, PRL, where he became a Professor and the Head in 2015. His research interests include medical imaging, image and audio processing, digital humanities, and interpretable machine learning and the use of known operators.
Dr. Maier has been a member of the Steering Committee of the European Time Machine Consortium since 2016. In 2018, he has received the ERC Synergy Grant “4D nanoscope.”
\end{IEEEbiography}

\begin{IEEEbiography}[{\includegraphics[width=1in,height=1.25in,clip,keepaspectratio]{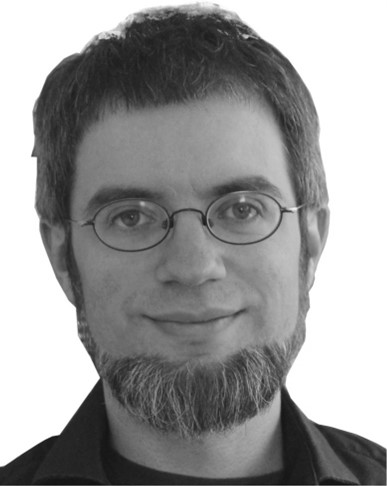}}]{Vincent Christlein}
received the degree in computer science and the Ph.D. (Dr.-Ing.) degree from FAU, Erlangen, Germany, in 2012 and 2018, respectively.
During his studies, he worked on automatic handwriting analysis with a focus on writer identification and writer retrieval. Since 2018, he has been a Research Associate with the PRL, FAU, where he was promoted to an Academic Councilor in 2020 and heads the Computer Vision Group, which covers a wide variance of topics, e.g., environmental projects such as glacier segmentation or solar cell crack recognition, but also computational humanities topics, such as document and art analysis.
\end{IEEEbiography}

\vspace{11pt}
\vfill

\setlength{\mathindent}{0cm}

\title{Supplementary Material -- SSL4SAR: Self-Supervised Learning for Glacier Calving Front Extraction from SAR Imagery}

\author{Nora Gourmelon$^{1, *}$, Marcel Dreier$^1$, Martin Mayr$^2$, Thorsten Seehaus$^3$, Dakota Pyles$^3$, Matthias Braun$^3$, \\Andreas Maier$^1$, Vincent Christlein$^1$
}

\maketitle

\section{SSL4SAR Dataset}
\begin{figure}[h]
    \centering
    \includegraphics[width=\linewidth]{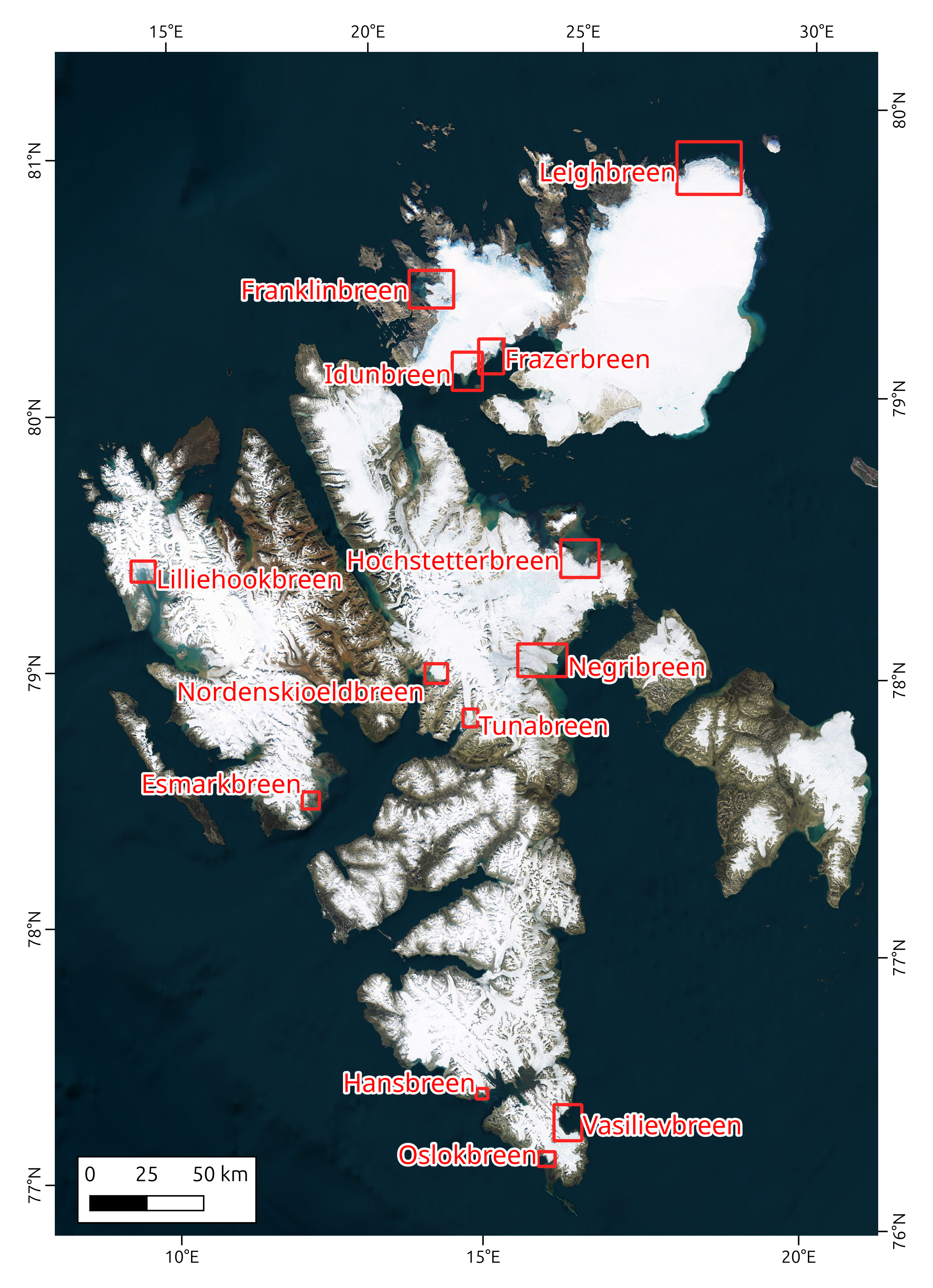}
    \caption{Map depicting the thirteen glaciers from Svalbard included in the SSL4SAR dataset. The \fboxsep=1pt\colorbox{red!100}{\color{white}red} boxes depict the bounding boxes used for cropping. Background: Bing Satellite {\textcopyright} Microsoft.}
    \label{fig:map}
\end{figure}
Figure~\ref{fig:map} illustrates the 13 glaciers on Svalbard included in the SSL4SAR dataset, with the 14th glacier being the Columbia Glacier in Alaska. These glaciers were selected to ensure a broad representation of glacier geometries, sizes, and orientations, with the goal of encouraging model generalization.

\section{Statistical Analysis of the Results}
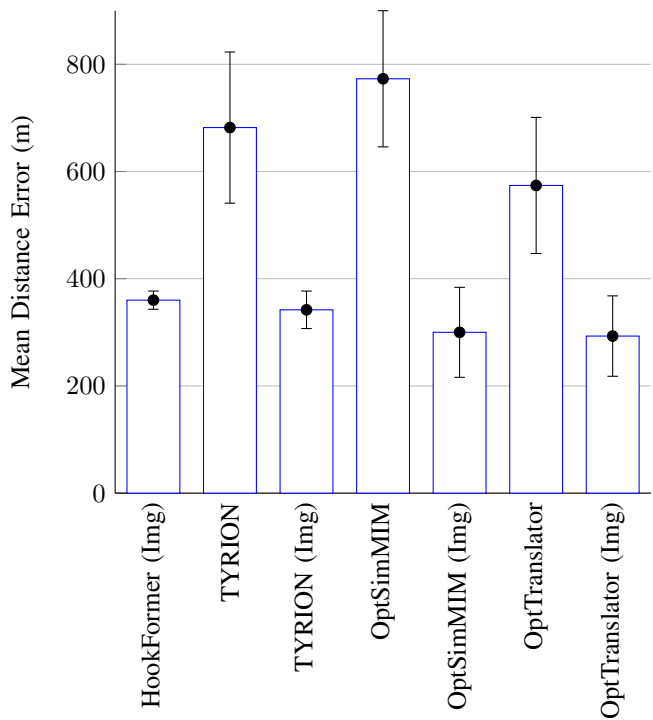
\begin{figure}[htbp]
    \centering
    \begin{tikzpicture}[scale=1.0, declare function={
            scale = 0.03782885669;
            a = 0.01;
            b = e^(a * (ln(3) + ln(41))/100);
            colorcalc(\nofront) = a * ln(\nofront + 1) / ln(b);
      }],
        \centering
        \begin{axis}[
        legend style={
              at={(1,0)},yshift=165pt,xshift=1em,anchor=north east,
              my area legend,
              name=legend,
              draw={none}, 
              fill={none},
            },
        legend image code/.code={%
            \draw[#1, draw=none] (0cm,-0.08cm) rectangle (0.4cm,0.08cm);
        },  
        height=8cm,
        width=0.48\textwidth,  
        bar width=20,
        ymax=900,
        ymin=0,
        xmin=0.5,
        xmax=7.5,
        ymajorgrids=true,
        axis y line*=left,
        axis x line*=bottom,
        xticklabels={HookFormer (Img), TYRION, TYRION (Img), OptSimMIM, OptSimMIM (Img), OptTranslator, OptTranslator (Img)},
        xtick={1, 2, 3, 4, 5, 6, 7},
        x tick label style={rotate=90,anchor=east},
        ylabel={Mean Distance Error (m)}]
            \foreach \x/\mean/\nofront [evaluate=\nofront as \color using {colorcalc(\nofront)}] in {1/360/0, 2/682/0, 3/342/0, 4/773/0, 5/300/0, 6/574/0, 7/293/0} {
                \edef\temp{\noexpand\addplot[ybar,blue,forget plot,fill=blue!\color] coordinates {(\x,\mean)};
                }
                \temp
            }
            
            \addplot+[only marks,black,mark options={draw=black,fill=black},forget plot][error bars/.cd,y dir=both, y explicit]
            coordinates {
            
            (1,360) +- (0,17)
            (2,682) +- (0,141)
            (3,342) +- (0,35)
            (4,773) +- (0,127)
            (5,300) +- (0,84)
            (6,574) +- (0,127)
            (7,293) +- (0,75)
            };            
        \end{axis}
    \end{tikzpicture}%
    \caption{Overview of the \acp{mde} with confidence intervals for the HookFormer~\cite{Wu.2023_2}, TYRION and TYRION with the two pretraining approaches (OptSimMIM and OptTranslator). \emph{(Img)} indicates that the model was initialized with ImageNet pretrained weights.}
    \label{fig:confidence_all}
\end{figure}

\begin{table*}[htbp]
    \centering
    \caption{Results of the null hypothesis tests. The alternative hypothesis is A$>$B, i.e., the \ac{mde} of A is greater than that of B.  One-sided Mann-Whitney U-tests are used and the $\alpha$-levels are Bonferroni-corrected. $U$ gives the test statistic. Column $d$ provides the standardized effect size Cohen's $d$.}
    \label{tab:nhst}
    \begin{tabular}{p{0.09\textwidth} p{0.09\textwidth} p{0.09\textwidth}
    p{0.09\textwidth} p{0.09\textwidth} p{0.09\textwidth} p{0.05\textwidth} p{0.05\textwidth} p{0.05\textwidth} p{0.05\textwidth} } 
        \toprule
        \multicolumn{3}{c}{A} & \multicolumn{3}{c}{B} & $U$ & $p$-value & $alpha$ & $d$\\
        \emph{Architecture} &\emph{pretraining} &\emph{Weights} & \emph{Architecture} &\emph{pretraining} &\emph{Weights} & & & & \\
        \midrule
        TYRION & / & ImageNet & HookFormer & / & ImageNet & 18.0 & 0.15 & 0.008 & 1.44 \\
        TYRION & / & ImageNet & TYRION & OptSimMIM & ImageNet & 20.0 & 0.08 & 0.008 & 1.67 \\
        TYRION & / & ImageNet & TYRION & OptTranslator & ImageNet & 20.0 & 0.08 & 0.008 & 1.96 \\
        HookFormer & / & ImageNet & TYRION & OptSimMIM & ImageNet & 20.0 & 0.08 & 0.008 & 4.78 \\
        HookFormer & / & ImageNet & TYRION & OptTranslator & ImageNet & 20.0 & 0.08 & 0.008 & 5.36 \\
        TYRION & OptSimMIM & ImageNet & TYRION & OptTranslator & ImageNet & 15.0 & 0.35 & 0.008 & 0.13\\
        \bottomrule
        &&&&&&&&&\\
    \end{tabular}
\end{table*}

Figure~\ref{fig:confidence_all} shows the averaged \acfp{mde} of the single model experiments along with the corresponding confidence intervals.
Six one-sided Mann-Whitney null hypothesis tests~\cite{Field.2018} were performed to investigate the differences in model performance. 
The results are given in tab.~\ref{tab:nhst}.
The tests yielded non-significant results at the Bonferroni-corrected~\cite{Bortz.2010} alpha levels, which does not imply that no difference exists between the models.
A difference between the groups simply cannot be statistically proven with the available data~\cite{Field.2018}.
$p$-values are affected by sample size~\cite{Field.2018}, i.\,e., the number of times each experiment was repeated, which in this case would be five.
The small sample size reduces the power of the test statistics to deem a given effect significant~\cite{Field.2018, Bortz.2010}.
In fact, given the confidence intervals provided, TYRION pretrained with OptTranslator and initialized with ImageNet weights would only significantly (at the Bonferroni-corrected $\alpha$-level) outperform the HookFormer if each experiment showed an improvement of at least about \SI{92}{\metre} over the HookFormer instead of the \SI{67}{\metre}~\cite{Field.2018}.
This would mean that the result would only be significant if the model could already achieve an improvement of 26\,\% over the HookFormer.\\
However, and much more importantly, the calculated, non-standardized effect sizes~\cite{Bortz.2010} and the standardized effect sizes~\cite{Bortz.2010} (Cohen's $d$~\cite{Field.2018}) are large~\cite{Bortz.2010}. 
Effect sizes allow us to interpret the importance of an effect~\cite{Field.2018}, which in our case would be the superiority of one model over the other in terms of the \ac{mde}.
A large effect size indicates that the effect is substantial and important.
Finally, we argue that a type \uproman{1} error~\cite{Field.2018} (falsely believing that an effect -- a difference -- exists) would be less detrimental than a type \uproman{2} error~\cite{Field.2018} (falsely believing that no effect -- no difference -- exists) when effect sizes are large.

\section{Image-wise Results of the Ensemble Approaches}
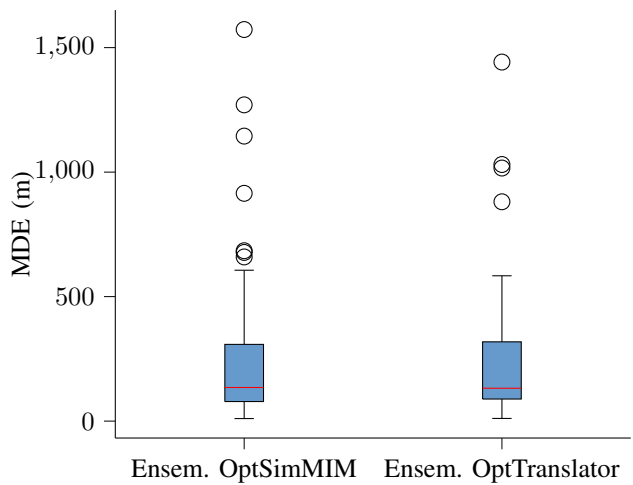
\begin{figure}[H]
    \centering
    \begin{tikzpicture}
        \begin{axis}[
        tick align=outside,
        tick pos=left,
        x grid style={black},
        xmin=0.5, xmax=2.5,
        xtick style={color=black},
        xtick={1,2},
        xticklabels={Ensem. OptSimMIM,Ensem. OptTranslator},
        y grid style={black},
        ylabel={MDE (m)},
        ymin=-67.9540516539048, ymax=1650.7775920384,
        ytick style={color=black},
        axis y line*=left,
        axis x line*=bottom,
        ]
            \addplot [black, fill=bluegray, fill opacity=1.0]
            table {%
            0.925 78.5704553373397
            1.075 78.5704553373397
            1.075 308.172963694856
            0.925 308.172963694856
            0.925 78.5704553373397
            };
            \addplot [black]
            table {%
            1 78.5704553373397
            1 10.1701139684726
            };
            \addplot [black]
            table {%
            1 308.172963694856
            1 605.980456197482
            };
            \addplot [black]
            table {%
            0.9625 10.1701139684726
            1.0375 10.1701139684726
            };
            \addplot [black]
            table {%
            0.9625 605.980456197482
            1.0375 605.980456197482
            };
            \addplot [black, mark=o, mark size=3, mark options={solid,fill opacity=0}, only marks]
            table {%
            1 677.435233005314
            1 659.328274525705
            1 1144.6913914643
            1 684.243745338616
            1 1270.13960585103
            1 1572.65342641602
            1 914.871756770948
            };
            \addplot [black, fill=bluegray, fill opacity=1.0]
            table {%
            1.925 88.6426408669911
            2.075 88.6426408669911
            2.075 318.423579000392
            1.925 318.423579000392
            1.925 88.6426408669911
            };
            \addplot [black]
            table {%
            2 88.6426408669911
            2 10.7567810470982
            };
            \addplot [black]
            table {%
            2 318.423579000392
            2 583.690371208578
            };
            \addplot [black]
            table {%
            1.9625 10.7567810470982
            2.0375 10.7567810470982
            };
            \addplot [black]
            table {%
            1.9625 583.690371208578
            2.0375 583.690371208578
            };
            \addplot [black, mark=o, mark size=3, mark options={solid,fill opacity=0}, only marks]
            table {%
            2 1016.36510556495
            2 1030.60761405626
            2 880.872357685498
            2 1442.02543613836
            };
            \addplot [red]
            table {%
            0.925 135.021363225215
            1.075 135.021363225215
            };
            \addplot [red]
            table {%
            1.925 132.054456705946
            2.075 132.054456705946
            };
        \end{axis}
    \end{tikzpicture}
    \caption{Overview of the \acp{mde} for the two in-domain pretrained ensemble models obtained using Setup 4 combined with the ensemble approach. Outliers are visualized as circles.}
    \label{fig:image_based_mde}
\end{figure}

Fig.~\ref{fig:image_based_mde} shows the distribution of image-wise \acp{mde} for the two in-domain pretrained ensemble models obtained using Setup 4 combined with the ensemble approach. Apart from a few outliers, the errors are tightly clustered around the median. For the OptSimMIM ensemble model, 75\,\% of \ac{mde} values lie below \SI{308}{\metre}; for the OptTranslator model, below \SI{318}{\metre}. 

\begin{figure*}
    \centering
    \includegraphics[width=0.75\linewidth]{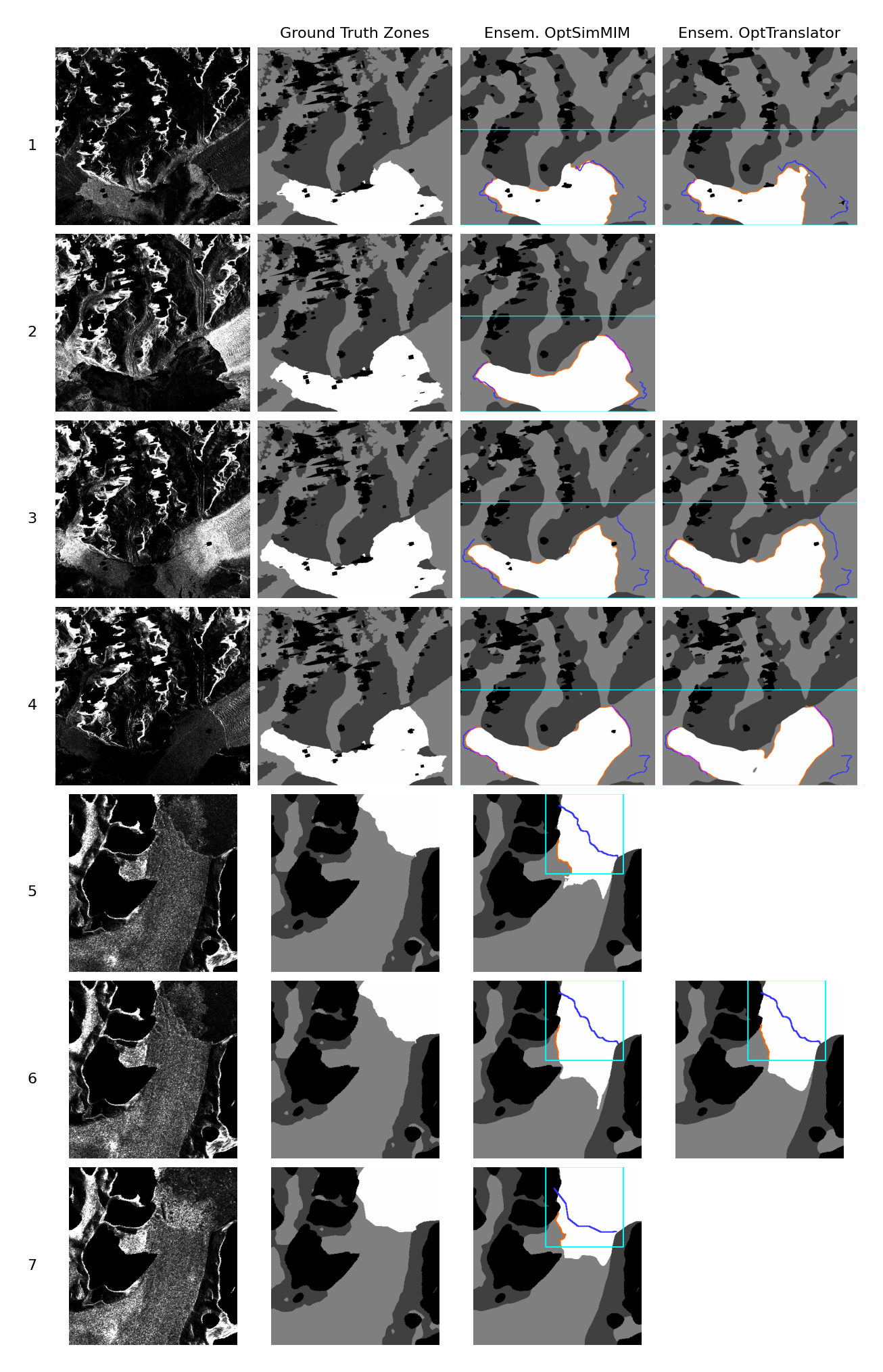}
    \caption{Visualization of the outliers of Fig.~\ref{fig:image_based_mde}. The \ac{sar} images are from 
    (1) Columbia Glacier taken by Sentinel-1 on 10 May 2016, 
    (2) Columbia Glacier taken by Sentinel-1 on 11 January 2017, 
    (3) Columbia Glacier taken by Sentinel-1 on 6 January 2018, 
    (4) Columbia Glacier taken by Sentinel-1 on 19 March 2018, 
    (5) Mapple Glacier taken by ENVISAT on 22 September 2007, 
    (6) Mapple Glacier taken by ENVISAT on 27 October 2007, 
    (7) Mapple Glacier taken by ENVISAT on 3 July 2010.
    If a prediction for an image is not an outlier the space in that row is left blank. The classes are visualized in white (ocean), light grey (glacier), dark grey (rock outcrop), and black (areas with no information).
    \fboxsep=1pt\colorbox{lava!100}{Orange} represents the predicted front, \fboxsep=1pt\colorbox{blue!100}{\color{white}blue} is used for the ground truth front, and \fboxsep=1pt\colorbox{magenta!100}{\color{white}pink} represents a perfect match between prediction and ground truth. The bounding box used during post-processing is given by \fboxsep=1pt\colorbox{Turquoise!100}{turquoise}. Columbia Glacier images are cropped for visualization purposes. The \ac{sar} images are shown without contrast enhancement to reflect the raw input provided to the model. Best viewed zoomed in.}
    \label{fig:outliers}
\end{figure*}

To investigate the causes of the outliers shown in Fig.~\ref{fig:outliers}, we examined individual cases in detail. All three Mapple Glacier images among the outliers display ice mélange in front of the calving front, which led both the OptTranslator and OptSimMIM ensemble models to misclassify parts of the glacier as melange. 
Conversely, in one Columbia Glacier outlier (\ac{sar} image 3 in Fig.~\ref{fig:outliers}), the reverse occurred: portions of the ice mélange were misclassified as glacial ice.
Another reason for the high \ac{mde} values in some of the Columbia Glacier outlier cases (images 1, 3, and 4) was that the upper glacier arm on the right side of the image was fronted by ice mélange that extended towards the open sea, visually separating the ice-free ocean in front of the lower glacier arm. The models misclassified this disconnected ice-free ocean region as glacier, likely due to similarities in backscatter with wet snow. Additionally, Columbia Glacier outliers frequently show erroneous predictions of calving fronts along the coastline, substantially increasing the \ac{mde}.

\vfill

\end{document}